\RequirePackage{fix-cm}
\documentclass[twocolumn]{svjour3}          
\smartqed  
\usepackage{graphicx}
\usepackage{url}
\usepackage{booktabs}
\usepackage{graphicx,subfigure,array,rotating,amsmath}
\usepackage[table]{xcolor}

\newcommand{\degree}{\ensuremath{^\circ}}
\newcommand {\1}{\columnwidth}
\newcommand {\2}{.48\columnwidth}
\newcommand {\3}{.80\columnwidth}

\usepackage{cite}
\begin{document}

\title{Sensor Fusion of Camera, GPS and IMU using Fuzzy Adaptive Multiple Motion Models
}


\author{Erkan Bostanci \and Betul Bostanci \and Nadia Kanwal \and Adrian F. Clark
}


\institute{E. B. \at
              Ankara University, Computer Engineering Dept., Golbasi, Ankara, Turkey \\
              Tel.: +90-312-2033300/1767\\
              \email{ebostanci@ankara.edu.tr}           
           \and
           B. B. \at
              HAVELSAN Inc.
              \email{bbostanci@havelsan.com.tr}
         \and
         N. K. \at
			Lahore College for Women University, Lahore, Pakistan
			\email{nadia.kanwal@lcwu.edu.pk}			
         \and
         A. F. C. \at
          School of Computer Science and Electronic Engineering, University of Essex, UK
          \email{alien@essex.ac.uk}
}


\maketitle

\begin{abstract}
A tracking system that will be used for Augmented Reality (AR) applications has two main requirements: accuracy and frame rate. The first requirement is related to the performance of the pose estimation algorithm and how accurately the tracking system can find the position and orientation of the user in the environment. Accuracy problems of current tracking devices, considering that they are low-cost devices, cause static errors during this motion estimation process. The second requirement is related to dynamic errors (the end--to--end system delay; occurring because of the delay in estimating the motion of the user and displaying images based on this estimate. This paper investigates combining the vision-based estimates with measurements from other sensors, GPS and IMU, in order to improve the tracking accuracy in outdoor environments. The idea of using Fuzzy Adaptive Multiple Models (FAMM) was investigated using a novel fuzzy rule-based approach to decide on the model that results in improved accuracy and faster convergence for the fusion filter. Results show that the developed tracking system is more accurate than a conventional GPS--IMU fusion approach due to additional estimates from a camera and fuzzy motion models. The paper also presents an application in cultural heritage context.
\end{abstract}

\section{Introduction}
\label{sec:introduction}

Augmented Reality (AR) is the process of blending real-world images with artificial objects or information generated by the computer. It is defined as an extension of user's environment with synthetic content~\cite{Broll2008}. AR can also be used to enrich human perception and facilitate the understanding of complex 3D scenarios~\cite{Azuma1997, Ribo2002}.  

Tracking, the process of locating a user in an environment, is critical to the accuracy of AR applications as more realistic results are obtained in the presence of accurate AR registration~\cite{Bostanci2013a}. This process includes determining the position and orientation of the AR user. Generally, the most important part of this process is tracking the head, as the user typically wears a head mounted display on which the augmented images of the real world are displayed. Furthermore, a tracking system mounted on the head has better signal reception if GPS will be used, has a good field-of-view of the scene for visual tracking, and removes the need for lever-arm compensation (\emph{i.e.} adjusting the reference frames of the tracking system to align with the user's eyes).

Current user tracking systems still have many problems due to two types of errors were defined in~\cite{Azuma1997}, namely static and dynamic errors. The errors in tracking systems are considered as static errors due to the inadequate accuracy provided by current low-cost sensors. Dynamic errors are due to end-to-end system delay which is the time elapsed between the time when the tracking system measures the position and orientation of the user to the time when the images appear on the display. For the static errors mentioned above, there are a number of uncertainties related to the sensors used for obtaining motion estimates as measurements for the fusion filter. These uncertainties and imprecisions arise from the accuracy problems of the GPS, loss of fine motion detail in the vision-based approach and drift problems in case of the IMU. A fusion of multiple sensors and combining estimates from them significantly reduces these problems and provide more accurate results. The motion pattern followed by the user introduces another type of uncertainty as will be elaborated later in the paper. Referring to the dynamic errors, it is known that the human visual system can process 8--10 images in a second and current industrial standards for frame rate is between 25 and 30 frames per second~\cite{Read2000}, hence a tracking system should be fast enough to produce rendered results to be displayed to the user.

The improved accuracy of an AR system due to tracking also prevents problems such as visual capture~\cite{Azuma1997} and does not allow visual sensors to gain priority over other sensors. For instance, inadequate registration accuracy can cause the user to reach or walk to the wrong part of the real environment because the augmentation has been displayed on another part. The eyes of the users get used to the error in the virtual environment and after some time of usage they start to accept these errors as correct, which is not desirable.

Fuzzy logic makes use of the linguistic variables which are normally used to represent verbal uncertainties in order `to exploit the tolerance in cases of uncertainty and imprecision'~\cite{Zadeh1994}. In this context, the uncertainties arise from the user motion since this cannot be predicted by the tracking system and the imprecision is due to the static errors mentioned above since low-cost sensors are employed in the application.

The contributions of this paper are two-fold for user tracking systems for outdoor environments which is known to be more challenging than indoors that provide a structured environment for tracking~\cite{Bostanci2013a}. The first contribution is combining motion estimates from a camera, GPS and an IMU, with all sensors being low-cost. The approach here uses keyframes extracted from the camera and find the vision-based motion estimate. This estimate is applied to the readings from GPS and IMU sensors for obtaining the final pose in a Kalman filtering framework. Vision-based tracking~\cite{Bostanci2012c} is quite useful because images of the environment are already required for augmentation. With this available information, it is wise to use these images for both finding a motion estimate for the user and overlaying the synthetic models. The second contribution is the multiple motion models using a fuzzy rule-base in order to select the motion model that will dynamically adapt the motion pattern followed by the user eventually reducing the filter error. 

It will be shown that this is best achieved using a combination of position and orientation estimates from several sensors, using efficient algorithms to perform user tracking in real-time.

The rest of the paper is structured as follows: Section~\ref{sec:relatedWork} presents the literature review on fusion approach for user tracking and using multiple motion models for filtering. Section~\ref{sec:findingMotionEstimates} described the approaches used to obtain motion estimates from the different sensors used in this study followed by Section~\ref{sec:sensorFusion} where these estimates are combined in a Kalman filter. Section~\ref{sec:fuzzyMotionModels} present the idea of using fuzzy adaptive motion models. Results are presented in Section~\ref{sec:results} and a sample application is given in user tracking in Section~\ref{sec:application}. Finally, the paper is concluded in Section~\ref{sec:conclusions}.

\section{Related Work}
\label{sec:relatedWork}

Sensor fusion for tracking involves combining motion estimates from two or more sensors.  Integration of data from GPS and IMU  has been well-studied~\cite{Groves2008} in order to improve upon the robustness of the individual sensors against a number of problems related to accuracy or drift. The Kalman filter is known to be the most widely-used filter mainly due to its simplicity and computational efficiency~\cite{Kaplan2005} with some drawbacks related to linearity problems~\cite{Thrun2006}.  

The literature presents examples of combining GPS with vision rather than inertial sensing. For instance, Schleicher \emph{et al.}~\cite{Schleicher2009} used stereo cameras with a low-cost GPS receiver in order to perform vehicle localization with a submapping approach. Armesto \emph{et al.}~\cite{Armesto2007} used a fusion of vision and inertial sensors in order to perform pose estimation for an industrial robot by using the complementary characteristics of these sensors~\cite{Bostanci2013a}. GPS position was combined with visual landmarks (tracked in stereo) in order to obtain a global consistency in~\cite{Konolige2008b}. A similar approach was followed by Agrawal \emph{et al.}~\cite{Agrawal2007} on an expensive system using four computers. 

Visual-inertial tracking has also become a popular technique, due to the complementary characteristics of the sensors, and is used in many different applications~\cite{Chroust2004}. Vision allows estimation of the camera position directly from the images observed~\cite{You1999}. However, it is not robust against 3D transformations, and the computation is expensive. For inertial trackers, noise and calibration errors can result in an accumulation of position and orientation errors. It is known that inertial sensors have long term stability problems~\cite{Chroust2004}. Vision is good for small acceleration and velocity. When these sensors are used together, faster computation can be achieved with inertial sensors and the drift errors of the inertial sensor can be corrected using vision. Applications generally use low frequency vision data and high frequency inertial data~\cite{Chen2004} since visual processing is more expensive and trackers today can generate estimates at rates up to $550$Hz using custom hardware~\cite{Lang2002}.

Bleser~\cite{Bleser2009} combined vision-based motion estimates with IMU in a particle filtering framework for AR in indoor environments. In~\cite{Tornqvist2009}, a similar approach was followed in order to perform localization for a UAV: vision (a camera facing downwards) and inertial sensing was used together in a particle filter for position and orientation estimation in 2D. For estimating the altitude, a pressure sensor was used.

Recently, Oskiper \emph{et al.}~\cite{Oskiper2012} developed a tightly-coupled EKF visual--inertial tracking system for AR for outdoor environments using a relatively expensive sensor (XSens, MTi-G). The system used feature-level tracking in each frame and measurements from the GPS in order to reduce drift. In addition to this, a digital elevation map of the environment was used as well as a pre-built landmark database for tracking in indoor environments where GPS reception is not available (although it was claimed that no assumption about the environment was made). The error was found to be $1.16$ metres.

Attempts to improve the accuracy of the filtering have also been made using adaptive approaches. In some studies, values for the state and measurement covariance matrices were updated based on the innovation~\cite{Almagbile2010} and recently fuzzy logic was used for this task~\cite{Tseng2011,Kramer2012}. Another approach for fusing accelerometer and gyroscope for attitude estimation is also based on fuzzy rules~\cite{Ojeda2002} in order to decide which of the accelerometer or the gyroscope will be given weight for estimation based on observations from these sensors such as whether a mobile robot is rotating or not. A later approach~\cite{Hong2003} used the error and dynamic motion parameters in order to decide which sensor should have a dominant effect on the estimation.

Some other studies suggest~\cite{Chen2004} or use~\cite{Torr2002,Kanatani2004,Schindler2006,Civera2008b} the idea of employing different motion models for recognizing the type of the motion for two-view motion estimation and visual SLAM. Different studies~\cite{Torr2002,Kanatani2004,Schindler2006} used geometric two-view relations such as general, affine or homography in order to fit these models to a set of correspondences and using the outliers for obtaining a penalty score in a Bayesian framework. 

Civera \emph{et al.}~\cite{Civera2008b} used a bank of EKFs in order to apply different motion models to several filters concurrently and select the best model in a probabilistic framework. This approach incorporated 3 motion models, namely stationary, rotating and general, separating models for motions including translations and rotations.

The tracking system developed in this paper takes a different approach by combining estimates from a camera, a low-cost GPS and IMU sensors\footnote{\url{http://www.phidgets.com/}}  for better accuracy in motion estimation. A second novel contribution presented here is employing fuzzy logic to choose the best-fitting of several possible motion models, ensuring that the filter state is more consistent with the measurements and hence converges faster. Furthermore, the design here does not bring additional computational burden due to the simple design and efficient implementation of the rule-base.

\section{Finding Motion Estimates}
\label{sec:findingMotionEstimates}
Before describing the details of the sensor fusion algorithm, this section describes the methods used to obtain measurements from the camera, GPS and IMU.

\subsection{Vision-based approach}
A vision-based user tracking algorithm was presented in~\cite{Bostanci2012c} providing motion estimates obtained using a two-view approach, by calculating the essential matrix between the most recent two keyframes. The algorithm extracted a new keyframe based on the number of feature matched. The motion estimate between the keyframes was in form of a rotation ($R_x$, $R_y$ and $R_z$) and translation ($t_x$, $t_y$ and $t_z$) which were incorporated into a transformation matrix:

\begin{equation}
\resizebox{\hsize}{!}{$
Tr=
\left[
\begin{array}{cccc}
\cos R_y \cos R_z & -\cos R_y \sin R_z  & \sin R_y & t_x \\ 
\sin R_x \sin R_y \cos R_z +\cos R_x \sin R_z  & -\sin R_x \sin R_y \sin R_z + \cos R_x \cos R_z & -\sin R_x \cos R_y & t_y \\ 
-\cos R_x \sin R_y \cos R_z +\sin R_x \sin R_z  & \cos R_x \sin R_y \sin R_z +\sin R_x \cos R_z  & \cos R_x \cos R_y  & t_z \\ 
0 & 0 & 0 & 1
\end{array} 
\right]
$}
\label{eq:finalTransformation}
\end{equation}

This transformation matrix can be applied, following a dead-reckoning approach, to the last position estimated by the camera in order to obtain the new position when a camera is used as the sole sensor. In the sensor fusion algorithm, this transformation will be applied to the estimate obtained using GPS and IMU.

\subsection{GPS motion estimate}
The data obtained from the GPS is in well-known NMEA format and includes position, the number of visible satellites and detailed satellite information for a position $P$ on Earth's surface. Using this information, the GPS coordinates can be converted from geodetic latitude ($\phi$), longitude ($\lambda$) and altitude ($h$) notation to ECEF Cartesian coordinates  $\mathbf{x_{gps}}$, $\mathbf{y_{gps}}$ and $\mathbf{z_{gps}}$ as:
\begin{equation}
\begin{array}{l}
\displaystyle x_{gps} = (N+h)\cos(\phi)\cos(\lambda)\\
\displaystyle y_{gps} = (N+h)\cos(\phi)\sin(\lambda)\\
\displaystyle z_{gps} = ((1-e^2)N + h)\sin(\phi)
\end{array} 
\label{eq:latLongAlttoxyz}
\end{equation}
where
\begin{equation}
N = \frac{a}{\sqrt{1.0-e^2\sin(\phi)^2}}
\label{eq:N}
\end{equation}
and $a$ is the ellipsoid constant for equatorial earth radius (6,378,137m), $e^2$ corresponds to the eccentricity of the earth with a value of $6.69437999\times10^{-3}$~\cite{Kaplan2005}. The calculated values form the measurements from the GPS sensor as $m_{gps}=\left( x_{gps},y_{gps},z_{gps}\right)$.

\subsection{IMU motion estimate}
The IMU is used for both calculating a position estimate that will be combined with estimates from other sensors and generating the orientation estimate using a recent IMU filter by Madgwick~\cite{Madgwick2010}. Before finding these motion estimates from this sensor it is important to find noise parameters using a simple calibration stage described in the following.

\subsubsection{Sensor calibration}
\label{sec:ch6sensorCalibration}

The IMU used in the experiments is calibrated in the factory in order to prevent production-related problems such as sensor sensitivity and cross-axis misalignment. Nevertheless, the sensor generates non-zero values which are known as bias (offset) parameters at rest. Sensor calibration in this case simply consists of finding the values which are generated by the IMU while it is still.

It is performed by placing the IMU on a flat and stable surface (it was observed that even the vibrations from the computer can affect the parameters) and taking samples ($\sim5000$, which takes around 30 seconds). The samples for the accelerometer ($a_x$, $a_y$, $a_z$) and the gyroscope ($g_x$, $g_y$, $g_z$),  are accumulated and their mean is found as the bias for each axis for both sensors. These offsets, presented in Table~\ref{tab:imuCalibration}, are then subtracted from each reading to find the actual amount of acceleration or rate of turn.

\begin{table}[h!t]
  \centering
  \caption{Calibration parameters found for accelerometer and gyroscope}
    \begin{tabular}{rccc}
    \toprule
          & \multicolumn{3}{c}{\textbf{Offsets}} \\
    \midrule
          & \textbf{x} & \textbf{y} & \textbf{z} \\
    \multicolumn{1}{c}{\textbf{Accelerometer}} & -0.000817 & 0.158242 & 0.987314 \\
    \multicolumn{1}{c}{\textbf{Gyroscope}} & -0.216527 & -0.052387 & -0.183611 \\
    \bottomrule
    \end{tabular}
  \label{tab:imuCalibration}
\end{table}

In addition to finding the bias parameters and subtracting them from the readings, a second approach for reducing the noise is to accumulate a set of readings (\emph{e.g.} four samples) and using their mean in order to reduce the effect of noise in position and orientation estimates from the IMU. Use of these calibration parameters and averaging several readings reduced the drift to a mean of $0.5\degree$ per minute when these parameters were used with the IMU filter described below. 

\subsubsection{Position and orientation estimates}
\label{sec:ch7position and orientation estimates}

Finding the position estimate from the IMU is performed by double-integrating the accelerometer outputs for several samples, the current implementation uses four samples. The first integration, to find the velocity, involves integrating accelerations using $v(t)=v(0)+at$:
\begin{equation}
\begin{array}{l}
\displaystyle v_x=\int_{0}^{T} a_x \mathrm{d}t = v_x(T)-v_x(0)\\
\displaystyle v_y=\int_{0}^{T} a_y \mathrm{d}t = v_y(T)-v_y(0)\\
\displaystyle v_z=\int_{0}^{T} a_z \mathrm{d}t = v_z(T)-v_z(0)
\end{array} 
\label{eq:imuVelocity}
\end{equation}
Since multiple samples are taken, $\mathrm{d}t$ is the time passed for each one of them. The next step is to integrate the velocities from (\ref{eq:imuVelocity}) to find the position using $x(t)=x(0)+vt$ as 
\begin{equation}
\begin{array}{l}
\displaystyle x_{imu}=\int_{0}^{T} v_x \mathrm{d}t = p_x(T)-p_x(0)\\
\displaystyle y_{imu}=\int_{0}^{T} v_y \mathrm{d}t = p_y(T)-p_y(0)\\
\displaystyle z_{imu}=\int_{0}^{T} v_z \mathrm{d}t = p_z(T)-p_z(0)
\end{array} 
\label{eq:imuPosition}
\end{equation}
These calculated positions ($m_{imu}=\left(x_{imu}, y_{imu}, z_{imu}\right)$) are used as the measurements from the IMU, used in both combining estimates from other sensors for the fusion filter presented here and a conventional GPS--IMU sensor fusion developed for comparison--see Section~\ref{sec:results}. 

\vfill
\section{Sensor Fusion Algorithm}
\label{sec:sensorFusion}
The motion estimates obtained by the individual sensors presented in Section~\ref{sec:findingMotionEstimates} are prone to error due to problems related to accuracy and drift. It makes sense to combine measurements from several sensors in order to exploit their characteristics, which are complementary to each other~\cite{Bostanci2013a} in order to yield more accurate results.  For this reason a sensor fusion approach was followed using a Kalman filter, since it is most common for such applications~\cite{Kaplan2005}.

The following subsections elaborate on the fusion filter, describing how the motion estimates from the three sensors are combined in a tightly-coupled design, the approach making use of multiple threads for efficiency and finally the tracking system using the sensor fusion algorithm presented here.

\subsection{Fusion filter}
The filter designed for integration of three sensors consists of a state $x$ which includes positional data ($P=(P_x,P_y,P_z)^T$), linear velocities ($V=(V_x,V_y,V_z)^T$), rotational data ($R=(R_x,R_y,R_z)^T$) and angular velocities ($\Omega=(\Omega_x,\Omega_y,\Omega_z)^T$):
\begin{equation}
 x= {\left(  P,  V,  R, \Omega \right)}^{T}
 \label{eq:ch6x}
\end{equation}

A simple state consisting of 12 elements will facilitate obtaining a better performance in speed than one with a larger state. At each iteration, the predict--measure--update cycle of the Kalman filter~\cite{Reid2001} is executed in order to produce a single output from several sensors as the filter output.

In the first stage, \emph{i.e.} prediction, a transition matrix ($F$ of (\ref{eq:ch6F})) is applied to the state $x$ in order to obtain the predicted position:

\begin{equation}
F=\left[
\begin{array}{cccccccccccc}
1 & 0 & 0 & \Delta t & 0 & 0 & 0 & 0 & 0 & 0 & 0 & 0 \\ 
0 & 1 & 0 & 0 & \Delta t & 0 & 0 & 0 & 0 & 0 & 0 & 0 \\ 
0 & 0 & 1 & 0 & 0 & \Delta t & 0 & 0 & 0 & 0 & 0 & 0 \\ 
0 & 0 & 0 & 1 & 0 & 0 & 0 & 0 & 0 & 0 & 0 & 0 \\ 
0 & 0 & 0 & 0 & 1 & 0 & 0 & 0 & 0 & 0 & 0 & 0 \\ 
0 & 0 & 0 & 0 & 0 & 1 & 0 & 0 & 0 & 0 & 0 & 0 \\ 
0 & 0 & 0 & 0 & 0 & 0 & 1 & 0 & 0 & \Delta t & 0 & 0 \\ 
0 & 0 & 0 & 0 & 0 & 0 & 0 & 1 & 0 & 0 & \Delta t & 0 \\ 
0 & 0 & 0 & 0 & 0 & 0 & 0 & 0 & 1 & 0 & 0 & \Delta t \\ 
0 & 0 & 0 & 0 & 0 & 0 & 0 & 0 & 0 & 1 & 0 & 0 \\ 
0 & 0 & 0 & 0 & 0 & 0 & 0 & 0 & 0 & 0 & 1 & 0 \\ 
0 & 0 & 0 & 0 & 0 & 0 & 0 & 0 & 0 & 0 & 0 & 1
\end{array} 
\right]
\label{eq:ch6F}
\end{equation}
where $\Delta t$ is the time between two prediction stages which can be computed using a timer in seconds to allow processing of 4 keyframes, see Section~\ref{sec:multithreaded} for details. This initial version of the transition matrix is relatively simple --using Constant Motion Model (CMM) which will be elaborated later in the paper; however fuzzy rules, described in Section~\ref{sec:fuzzyMotionModels}, will be used to decide on the velocity coefficients that will update this transition matrix.

The majority of the operations required for integrating the motion estimates from the three sensors are performed in the second stage, where measurements are taken and provided to the filter so that it can update itself. This stage can be examined separately for the position and orientation estimates. For the latter, the output of the IMU filter ($m_{R}=\left(yaw, pitch, roll\right)$ provide the rotational measurements used to update $R$ in (\ref{eq:ch6x}).

The idea of combining the positional estimates from the camera, GPS and IMU is due to the fact that the GPS is a discrete-time position sensor~\cite{Kaplan2005}. In order for the AR system used in application presented in Section~\ref{sec:application} to update the position of the virtual camera more rapidly, the position estimates from the fusion filter need to have smooth transitions between them in order to provide the impression of continuous motion. This is achieved by applying the transformation $Tr$, obtained from the motion estimate of the camera, to the position provided by the GPS sensor ($m_{gps}$) and then adding the motion estimate of the IMU ($m_{imu}$) as an offset: 
\begin{equation}
m_{P} = Tr \times m_{gps} + m_{imu}
\label{eq:ch6measurements}
\end{equation}
where $m_{P}$ constitutes the positional measurements for the fusion filter.

Having all the measurements ($m_{P}$ and $m_{R}$) ready, the filter can now update itself. After the update, the obtained estimates can directly be used by the AR system in order to update the position of the virtual camera for the application demonstrated in Section~\ref{sec:application}.

\subsection{Multi-threaded approach}
\label{sec:multithreaded}
The sensors used in the system had different data rates for delivering data and performing calculations to produce a motion estimate. These frequency differences resulted in a challenge while combining them to generate a single output for the AR system. To elaborate, the vision-based system can produce up to $4$ motion estimates per second while the IMU can produce up to $250$ samples per second which are to be used for the orientation filter and GPS can produce only a single measurement every second. Furthermore, AR processing has to produce a minimum of 10 frames per second in order to generate a smooth flow of the display.

A second challenge is due to the execution method of the GPS and the IMU sensors. The library handling these sensors was designed to be event-driven (\emph{i.e.} an event is triggered each time a datum is available.). The library makes an automatic call the related event handler when data from any of these two sensors become ready. This did not allow the handling of these sensors in the same thread, where other computations are performed in a procedural manner.

\begin{figure}
  \begin{center}
    \includegraphics[width=\columnwidth]{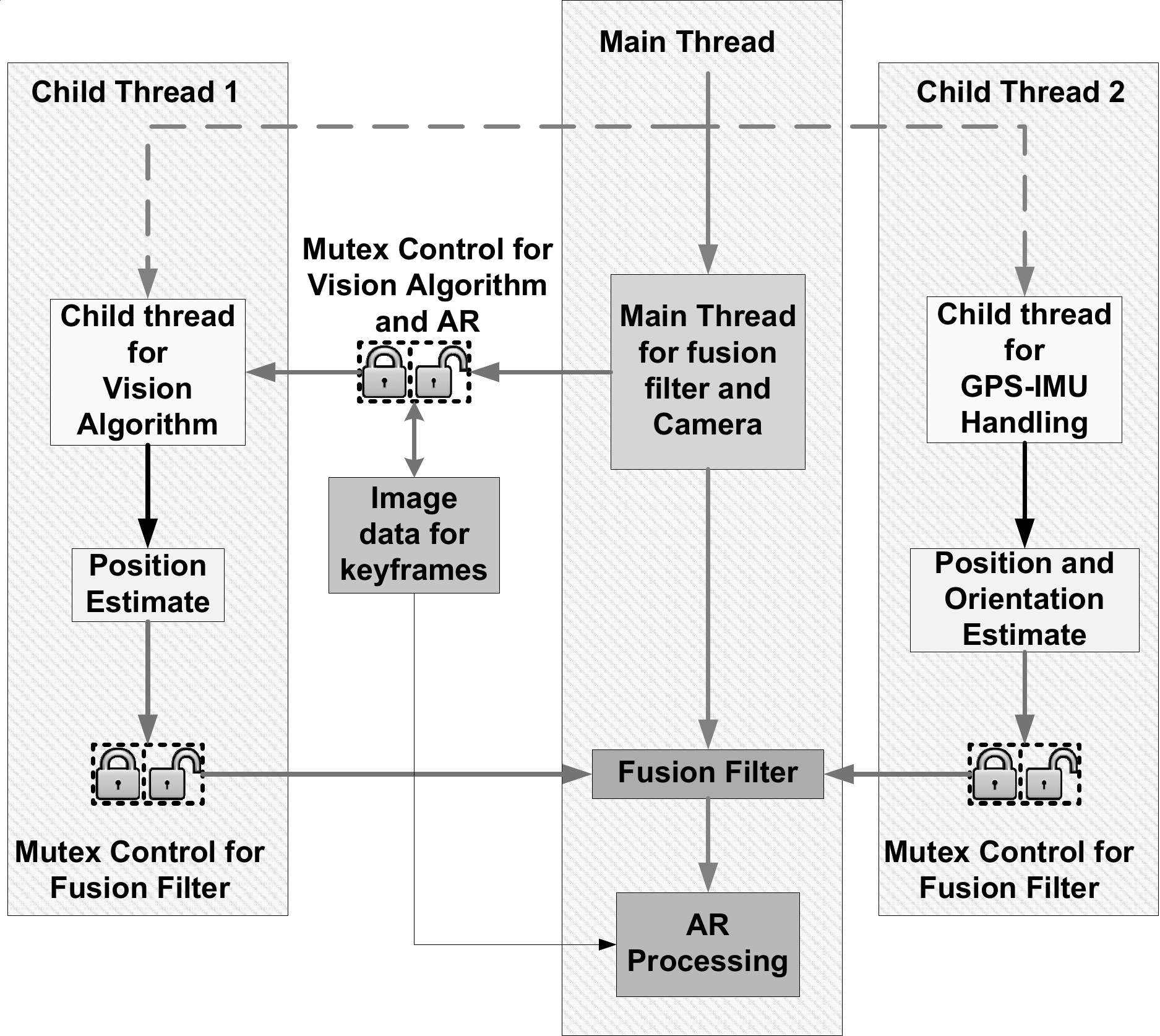}
  \end{center}
  \caption[Diagram of using threads to access data from different sensors]{Diagram of using multiple threads to access data from different sensors. The main thread is responsible for acquiring camera images, the fusion filter and using these two generating the AR output. The GPS and IMU handler is handled by a child thread since it is working based on events generated by these two sensors. The second child thread is responsible for the vision algorithm. Race conditions are prevented using three mutexes (shown with locks).}
  \label{fig:threadsForSensors}
\end{figure}

Due to the differences in data rates and the application logic used for the sensors employed in the system, a multi-threaded approach was followed as shown in Figure~\ref{fig:threadsForSensors}. The design used two child threads in addition to the main thread in order to circumvent the challenges mentioned above. The main thread is used to acquire camera images and pass these to both of the vision-based method as keyframes and AR processing. A child thread is used to handle vision-based processing by accessing the camera images acquired. The algorithm generates an estimate which will be later used by the fusion filter in the main thread. A second child thread handles the GPS and IMU sensors and computes the estimate generated by them. 

\subsection{Tracking system}
\begin{figure}[h!t]
  \begin{center}
    \includegraphics[width=\columnwidth]{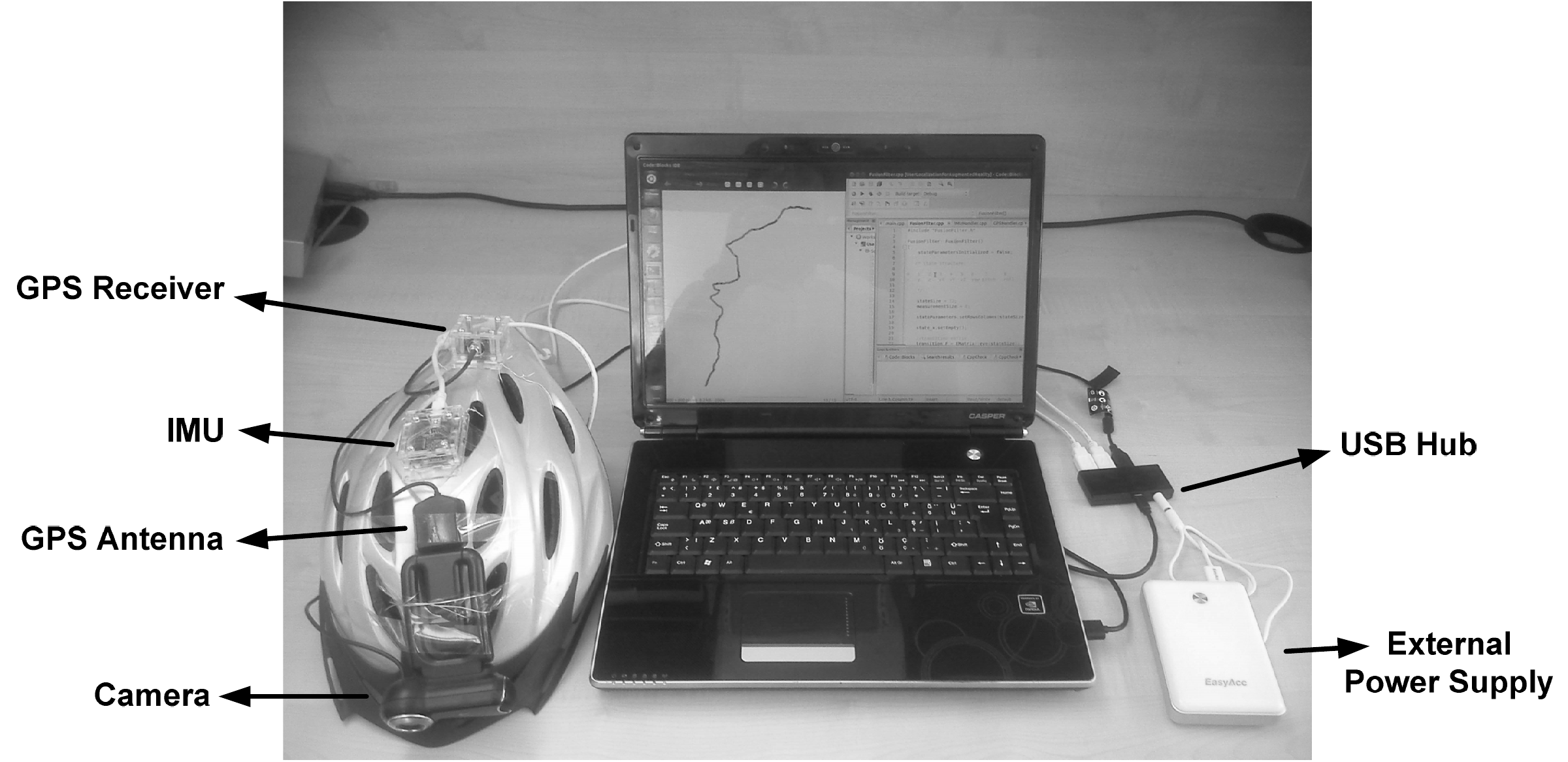}
  \end{center}
  \caption[Tracking system]{Tracking system}
  \label{fig:trackingSystem}
\end{figure}

The fusion algorithm was tested on a simple tracking system designed for this study. As shown in Figure~\ref{fig:trackingSystem}, the system consists of a laptop computer (Intel, dual core 2.80Ghz,  4GB memory with Linux operating system), a GPS receiver, an IMU and a web camera. An external power supply was also required for the hub connecting the sensors to the laptop since a single port was not able to provide enough power for the three sensors.

The placement of the sensors was chosen considering the lever-arm effect~\cite{Seo2005} which occurs particularly in GPS--IMU integration systems when sensors are placed apart from each other, so that the positional and rotational data sensed by them correspond to different positions. With this in mind, the camera, IMU and the antenna of the GPS are all placed 2--3 cm apart from each other, a negligible distance.

\section{Fuzzy Adaptive Multiple Motion Models}
\label{sec:fuzzyMotionModels}
Earlier in the introduction, it was mentioned that the motion patterns followed by the user causes additional uncertainty for the tracking system. A user, in a cultural heritage context, may follow a number of motion patterns, which may include stopping to examine ruins, walking slowly looking around or walking with a higher speed. The filter developed in the previous section uses a constant transition function ($F$ in (\ref{eq:ch6F})), or CMM, which does not take into account any information about the actual motion pattern performed by the user in the prediction stage. An improvement can be achieved if the filter is dynamically adapted based on the user's motion patterns. The reason behind this is that the dynamics of the filtering process is governed by both the internal parameters of the filter, such as the state $x$ and noise parameters ($Q$ for the process and $R$ for the measurement noise), and on the external side by the motion model used in $F$ for prediction. 

This section presents FAMM, a method of employing fuzzy logic to decide which one of the several motion models is the best fitting one, so that the filter state will be more in line with the measurements and hence converge faster.

\subsection{Handling the uncertainty in the fusion filter}
The idea presented here is to use adaptive motion models depending on the Kalman filter innovation ($y$) with an attempt to minimize the filter error. The innovation is actually hidden in the update part of the Kalman filter where the next value of the state is obtained using the measurements:
\begin{equation}
x_{i+1} = \hat{x}_{i+1}+K_{i}(z_{i}-h_{i}\hat{x}_{i+1})
\end{equation}

The difference between the measurements ($z$) and the prediction ($h\hat{x}$), omitting the subscripts indicating time, is defined as the innovation ($y$):
\begin{equation}
  y = z - h\hat{x}
  \label{eq:innovation}
\end{equation}
For the filter designed in the previous section, the innovation $y$ can be broken into two parts for positional ($y_p$) and rotational ($y_r$) data by taking the corresponding matrix elements. These elements refer to the measurement errors for positional and rotational estimates and will be used to decide on the motion model to be used in the next prediction stage with the aim of reducing filter error.

The design here makes use of nine motion models $MM$, each denoted as combinations of \texttt{P}$i$\texttt{R}$j$ where $i,j\in (0,1,2)$. Values for $i$ and $j$ are considered as velocity coefficients ($c_i$ and $c_j$) for the two components of the transition function ($F$) for position
\begin{equation}
\hat{x}_{P}=x_P+c_{i}V\Delta t
\end{equation}
and orientation as
\begin{equation}
\hat{x}_{R}=x_R+c_{j}\Omega\Delta t
\end{equation}

The idea presented here can be best described using examples of how these motion models work. For instance, \texttt{P0R0} indicates a stationary transition model where the current values of the state (\ref{eq:ch6x}) for position ($P$) and rotation ($R$) will be unchanged in the predicted state, whereas \texttt{P1R2} indicates a motion model where position is predicted with current positional velocities ($\hat{x}_{P}=x_P+(1 V)\Delta t$) but rotational velocities are doubled ($j=2$ so $\hat{x}_{R}=x_R+(2 \Omega)\Delta t$) to compensate for the effects of severe rotations.

\begin{figure}[h!t]
  \begin{center}
    \includegraphics[width=\1]{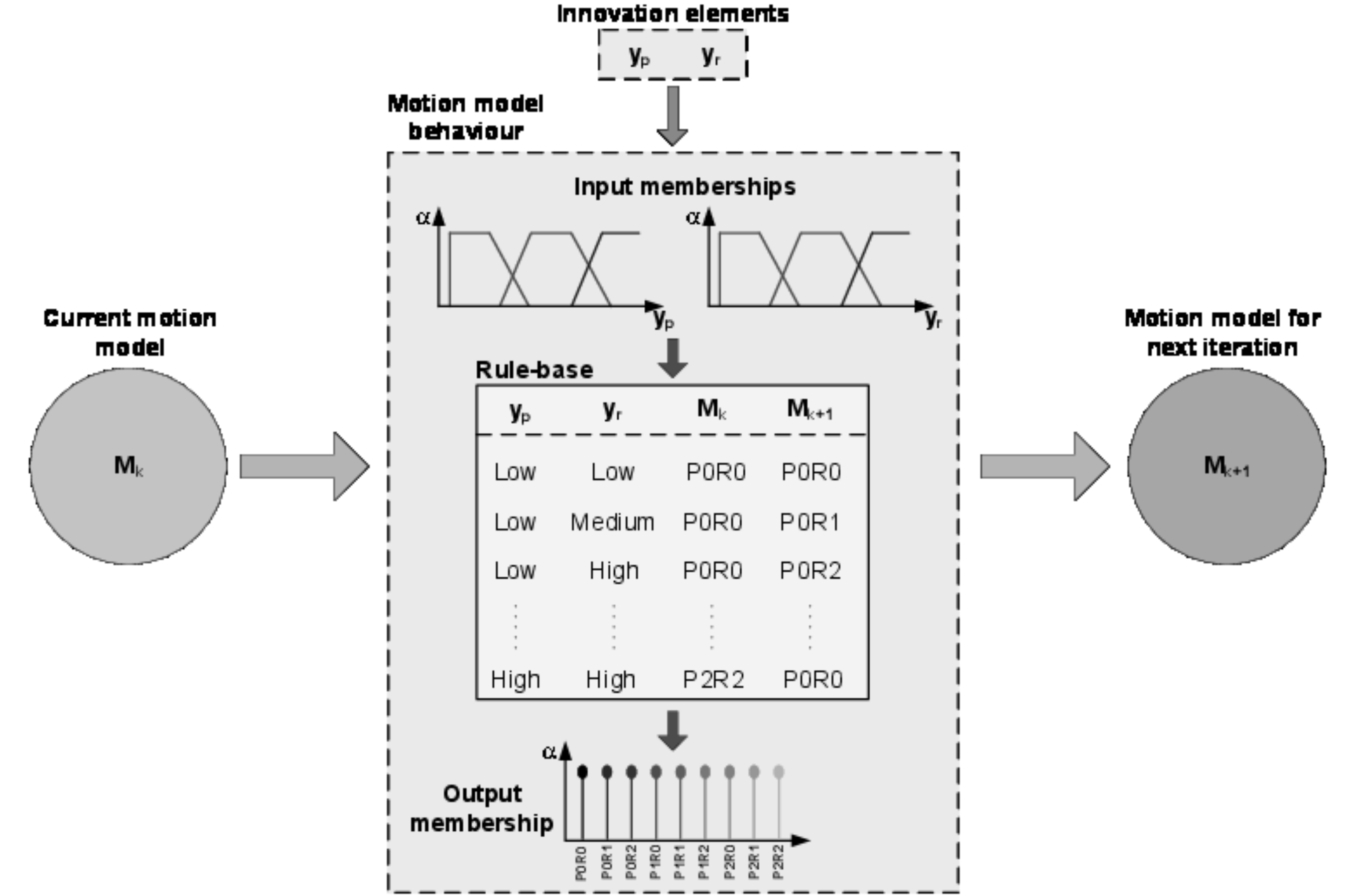}
  \end{center}
  \caption[Fuzzy rule-based selection of motion models]{Fuzzy rule-based selection of motion models. The Fuzzy Logic Controller (FLC) takes the two components of the innovation (\emph{i.e.} positional and rotational) and applies the membership functions in order to decide the firing strengths of the rules available in the rule-base. The antecedents of the rules are used to define the motion model used in the prediction stage of the next filter iteration.}
  \label{fig:ch6motionModelsActual}
\end{figure}

Actual selection of the motion model is achieved using a fuzzy logic controller which takes the $y_p$ and $y_r$ parameters, calculated in the update stage of the filter, and then applies a membership function to these parameters. In the implementation, magnitudes of $y_p$ and $y_r$ are calculated and then the input membership
functions are applied to them. The output of the membership function will define the `firing strengths' of rules. The rule with the maximum firing strength is selected to choose the motion model that will be used in the next prediction stage. This process is illustrated in Figure~\ref{fig:ch6motionModelsActual}.

Using the innovation values, a rule-base was designed for motion models which are used fill the components of the transition matrix ($F$) to provide this functionality.

\subsection{Rule-base definition}
The rule-base consists of the rules which can be examined in two parts namely the antecedent and the consequent. The antecedent part defines the conditions to be satisfied for the consequent to occur. In this case, the antecedents will include the membership values for the positional and rotational innovations and the current type of the motion model in order to select the motion model for the next prediction as the consequent.

A rule of the form \texttt{<Low, Medium, P0R0, P0R1>}, uses the first three components as antecedents and the last as the consequent, should be read as: 
\begin{quotation}
``\emph{IF} the positional innovation is \texttt{Low} \emph{AND} \\rotational innovation is \texttt{Medium} \emph{AND} \\the current motion model ($M_{k}$) is \texttt{P0R0}, \emph{THEN} \\change the motion model to \texttt{P0R1} ($M_{k+1}$) for the next iteration of the filter.''
\end{quotation}

The rule-base, a subset is presented in Table~\ref{tab:rulebase}, consists of $3^4=81$ rules $l^n$ where $l$ is the number of linguistic variables (three for \texttt{(Low, Medium,High)} and $n$ is the number of input variables (four for $y_p$, $y_r$ and $M_{k}$ which counts for two variables since $M_{k}=P_{i}R_{j}$). 

\begin{table}[h!t]
  \caption{Rule-base for motion models}
  {\begin{tabular}{cccc}\hline
      $y_p$ & $y_r$ & $M_k$ & $M_{k+1}$\\
      \hline
      Low & Low & P0R0 & P0R0 \\ 
      Low & Medium & P0R0 & P0R1 \\ 
      Low & High & P0R0 & P0R2 \\ 
      
      Low & Low & P0R2 & P0R2 \\ 
      Low & Medium & P0R2 & P0R1 \\ 
      Low & High & P0R2 & P0R0 \\ 
      
      Low & Low & P1R0 & P1R0 \\ 
      Low & Medium & P1R0 & P1R1 \\ 
      Low & High & P1R0 & P1R2 \\
      
      Low & Low & P1R2 & P1R2 \\ 
      Low & Medium & P1R2 & P1R1 \\ 
      Low & High & P1R2 & P1R0 \\
            
      Low & Low & P2R0 & P2R0 \\ 
      Low & Medium & P2R0 & P2R1 \\ 
      Low & High & P2R0 & P2R2 \\
            
      Low & Low & P2R2 & P2R2 \\ 
      Low & Medium & P2R2 & P2R1 \\ 
      Low & High & P2R2 & P2R0 \\
            
      Medium & Low & P0R0 & P1R0 \\ 
      Medium & Medium & P0R0 & P1R1 \\ 
      Medium & High & P0R0 & P1R2 \\ 
            
      Medium & Low & P0R2 & P1R2 \\ 
      Medium & Medium & P0R2 & P1R1 \\ 
      Medium & High & P0R2 & P1R0 \\ 
            
      Medium & Low & P1R0 & P2R0 \\ 
      Medium & Medium & P1R0 & P2R1 \\ 
      Medium & High & P1R0 & P2R2 \\ 
      
      Medium & Low & P1R2 & P2R2 \\ 
      Medium & Medium & P1R2 & P2R1 \\ 
      Medium & High & P1R2 & P2R0 \\
            
      Medium & Low & P2R0 & P1R0 \\ 
      Medium & Medium & P2R0 & P1R1 \\ 
      Medium & High & P2R0 & P1R2 \\ 
            
      Medium & Low & P2R2 & P1R2 \\ 
      Medium & Medium & P2R2 & P1R1 \\ 
      Medium & High & P2R2 & P1R0 \\
            
      High & Low & P0R0 & P2R0 \\ 
      High & Medium & P0R0 & P2R1 \\ 
      High & High & P0R0 & P2R2 \\ 
      
      High & Low & P0R2 & P2R2 \\ 
      High & Medium & P0R2 & P2R1 \\ 
      High & High & P0R2 & P2R0 \\ 
      
      High & Low & P1R0 & P0R0 \\ 
      High & Medium & P1R0 & P0R1 \\ 
      High & High & P1R0 & P0R2 \\ 
      
      High & Low & P1R2 & P0R2 \\ 
      High & Medium & P1R2 & P0R1 \\ 
      High & High & P1R2 & P0R0 \\ 
     
      High & Low & P2R0 & P0R0 \\ 
      High & Medium & P2R0 & P0R1 \\ 
      High & High & P2R0 & P0R2 \\ 
      
      High & Low & P2R2 & P0R2 \\ 
      High & Medium & P2R2 & P0R1 \\ 
      High & High & P2R2 & P0R0 \\
      \hline
    \end{tabular}}{}
  \label{tab:rulebase}
\end{table}

The rule-base presented here consists of a relatively large number of rules in order to handle all different transitions between motion models. For this reason, the rules are stored in a look-up table so that they can be accessed with a single query using the antecedent parameters. This design did not bring an extra computational overhead to the system, which is already using current resources at optimal capacity.

\subsection{Input/Output membership functions}
\begin{figure}[h!t]
  \begin{center}
    \subfigure[]{\includegraphics[width=\3]
      {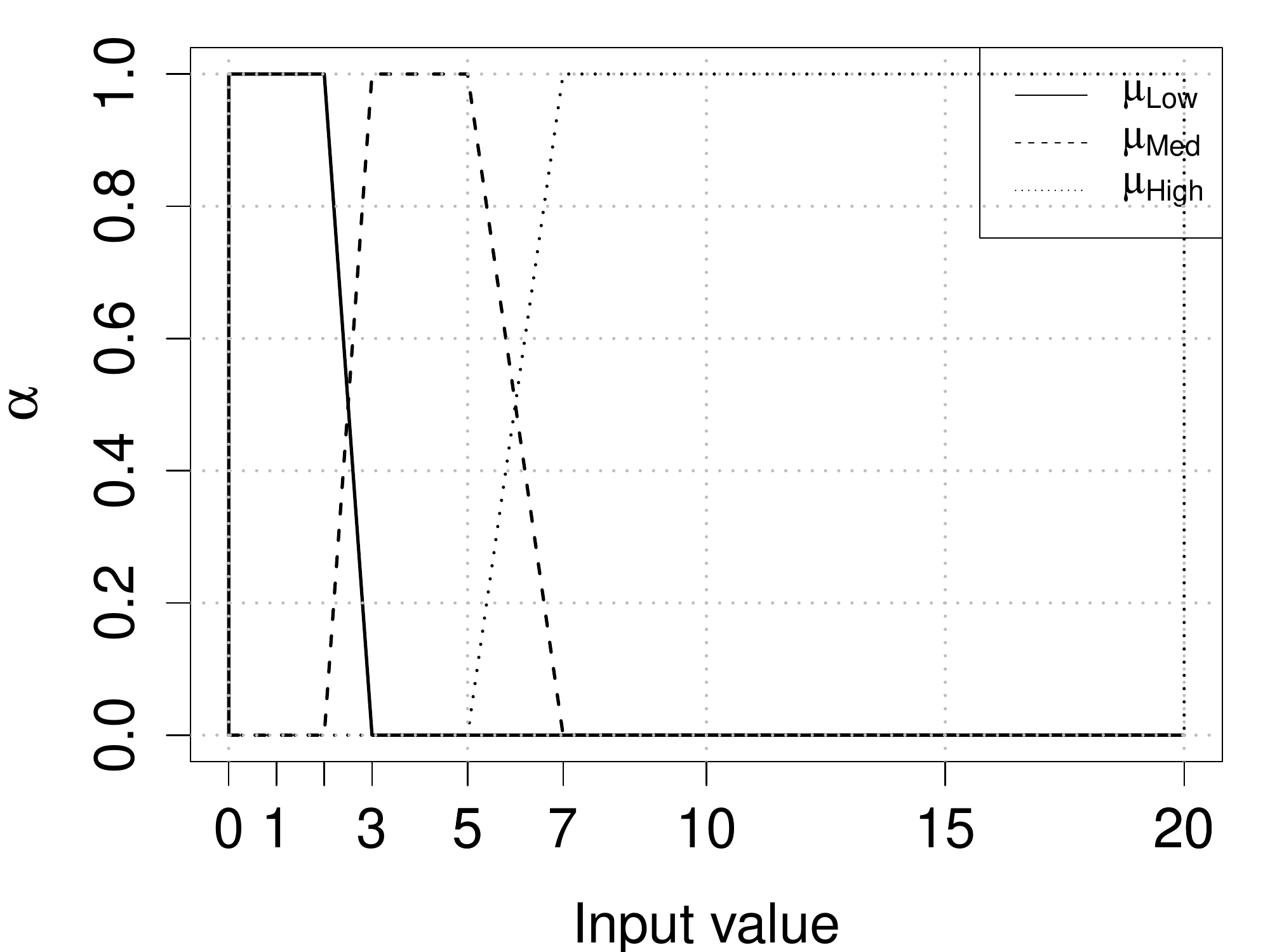}}
    \subfigure[]
    {\includegraphics[width=\3]{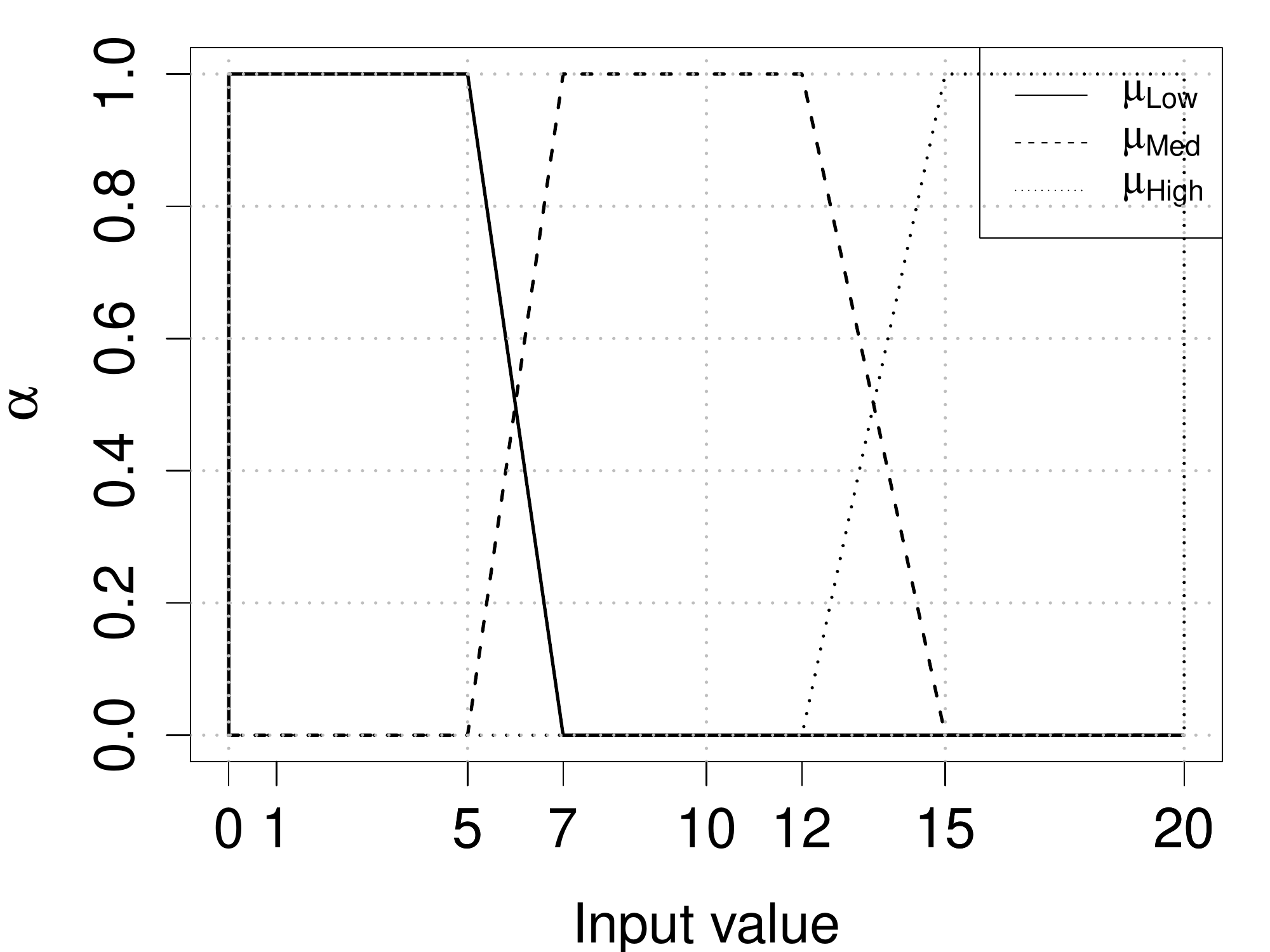}}
  \end{center}
  \caption{Input membership functions for positional and rotational innovations. (a) Positional innovation (b) Rotational innovation}
  \label{fig:inputMemberships}
\end{figure}

Earlier, it was mentioned that different velocity coefficients were used to allow a multiple motion models. These coefficients ($i, j \in (0,1,2)$) correspond to three fuzzy sets corresponding to three linguistic variables: \texttt{Low}, \texttt{Medium} and \texttt{High}. Calculation of the membership degrees for these linguistic variables are performed using the input membership functions defined as in Figure~\ref{fig:inputMemberships} using the crisp values of $y_p$ and $y_r$.

\begin{figure}[h!t]
  \begin{center}
    \includegraphics[width=\3]{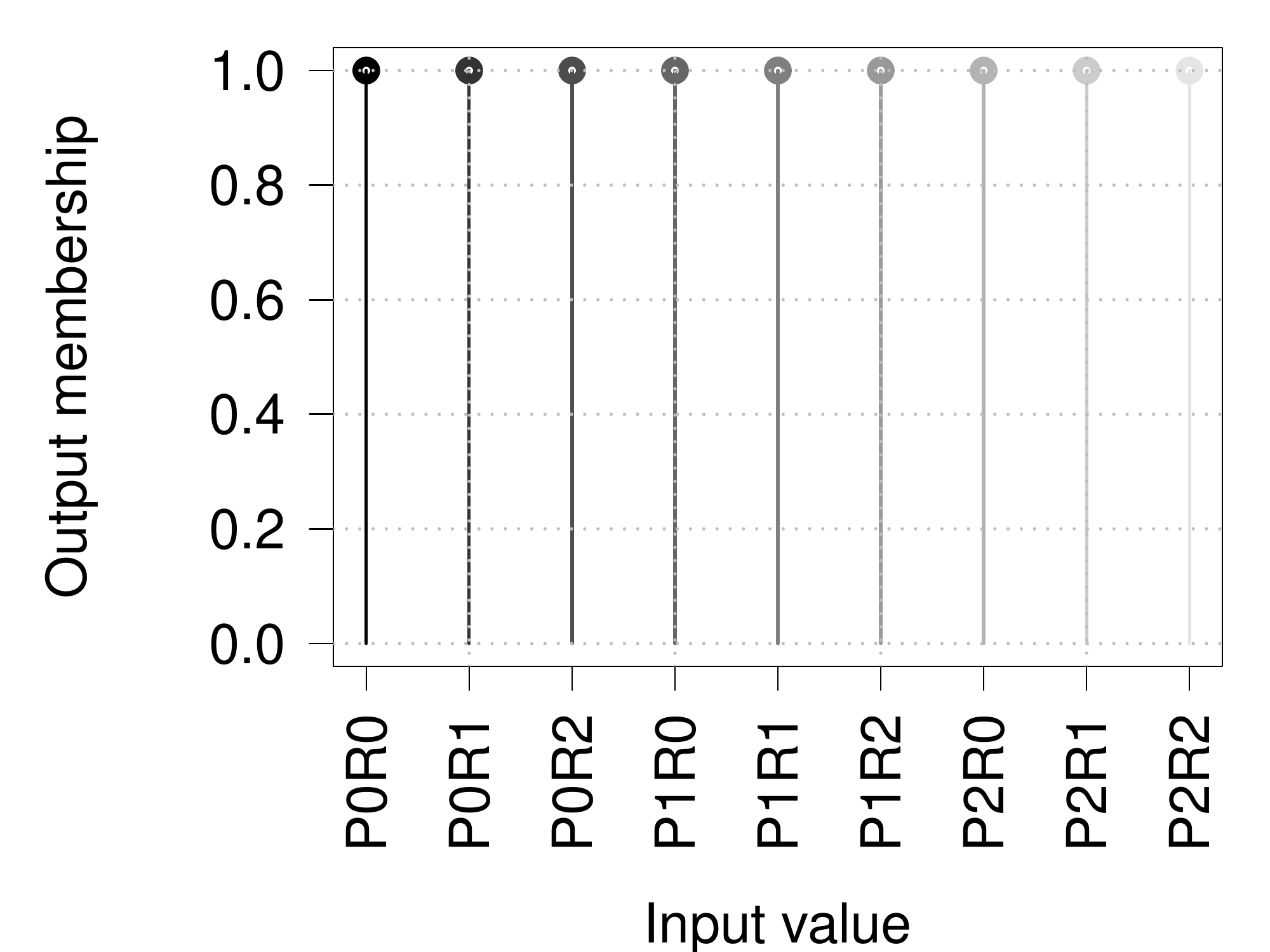}
  \end{center}
  \caption{Output membership function}
  \label{fig:outputMembershipFunctions}
\end{figure}

The output membership function of the FLC is a singleton suggesting only one type of motion model based on the results of the input membership function and the rule-base, as shown in Figure~\ref{fig:outputMembershipFunctions}. Note that the colours used to describe motion models will be used indicate the type of the motion model employed for different sections of the trajectory in Section~\ref{sec:results}.

\subsection{Processing}
The FLC is implemented as a motion model behaviour which receives the positional and rotational innovations are parameters. The \emph{AND} logical connector is represented using the product \texttt{t-norm}~\cite{Ross2009}. The firing strength of each is calculated by multiplying the membership values ($\mu_{Low}$, $\mu_{Med}$ and $\mu_{High}$) of positional and rotational innovations. The fire strengths for the rules which do not include the current motion model in its third antecedent are simply set to zero and the consequent of the rule with the maximum fire strength is selected as the motion model for the next prediction step.

\section{Results}
\label{sec:results}
\begin{figure}
  \begin{center}
    \subfigure[]{\includegraphics[width=\2]{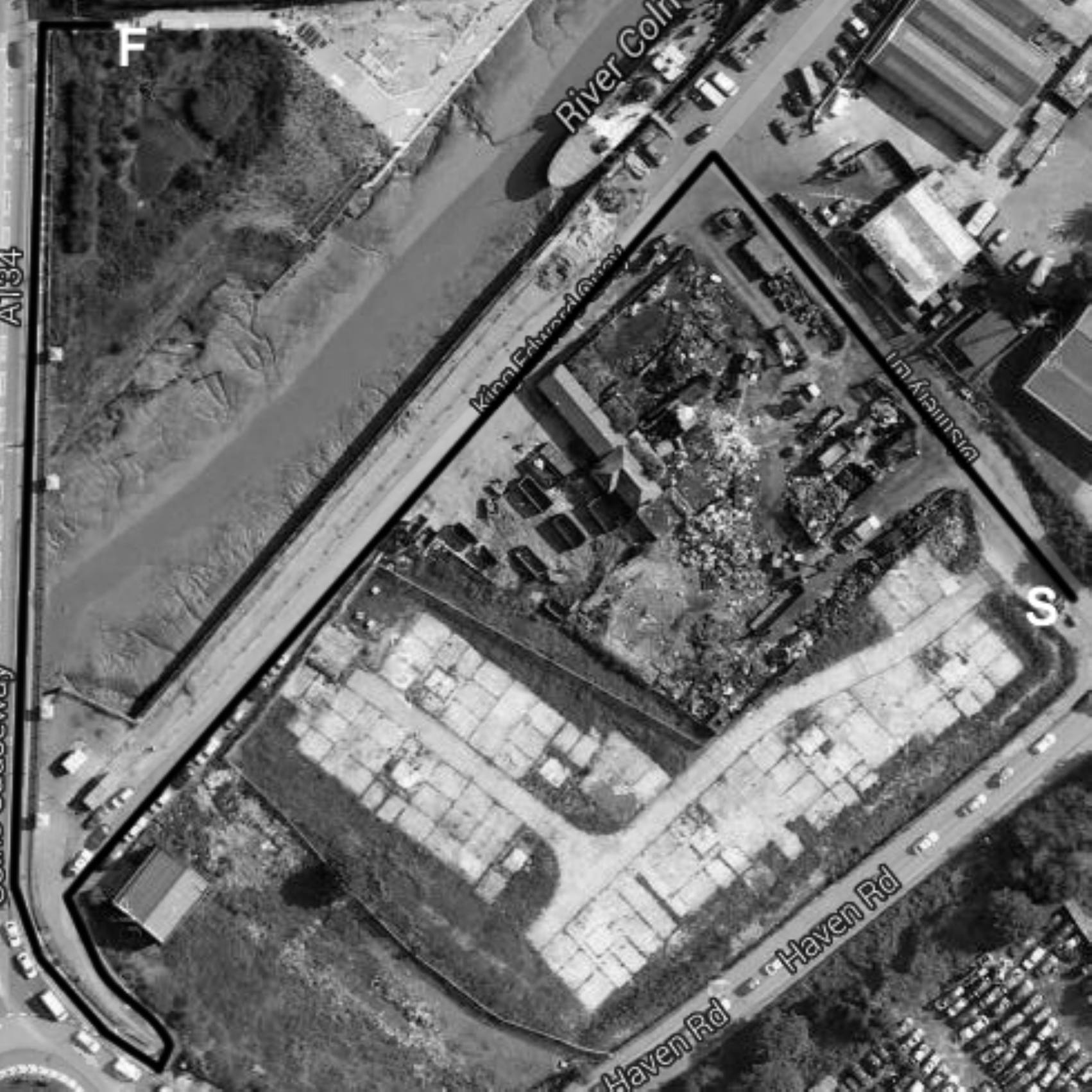}}\\
    \subfigure[]{\includegraphics[width=\2]{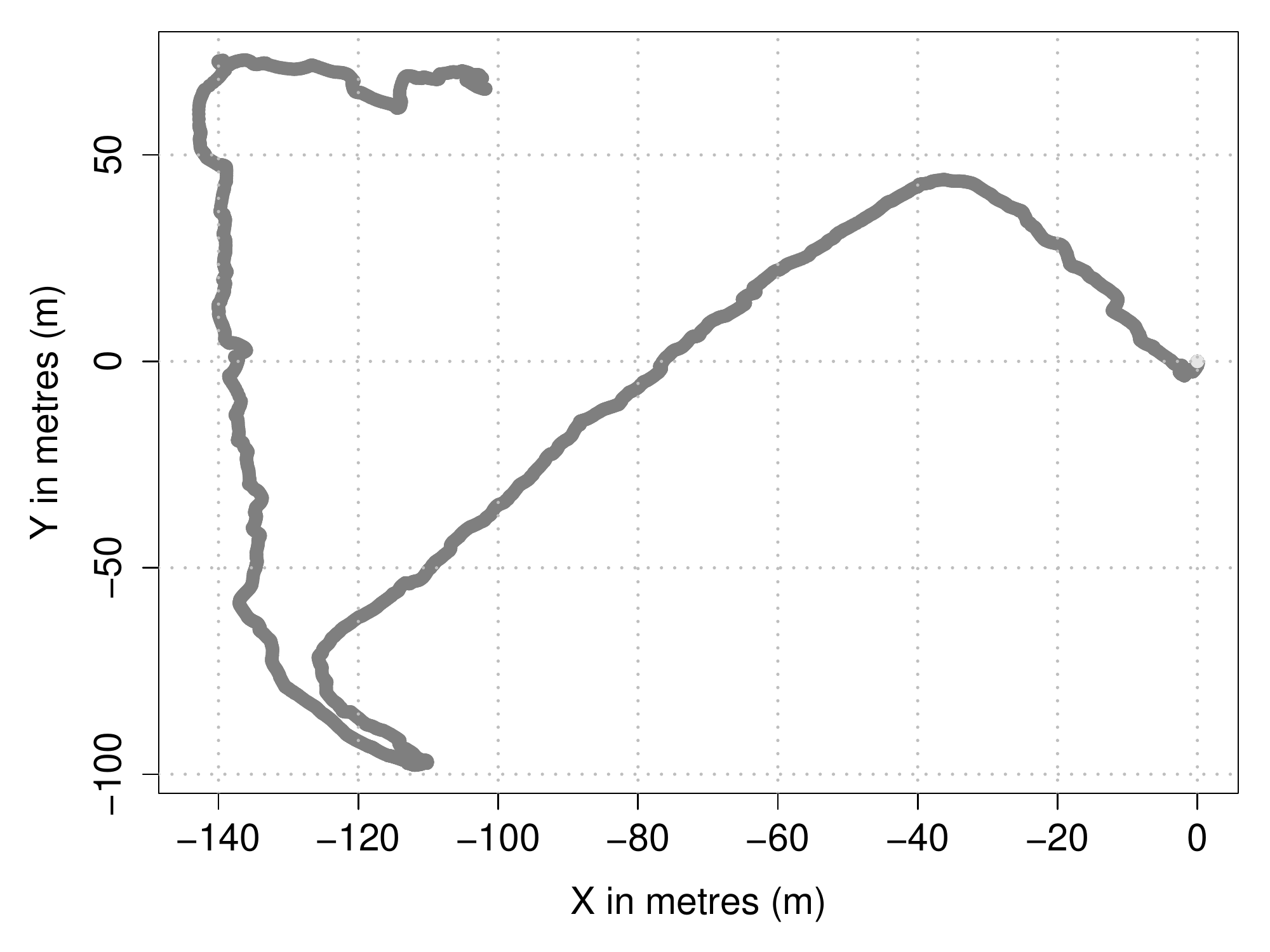}}
    \subfigure[]{\includegraphics[width=\2]{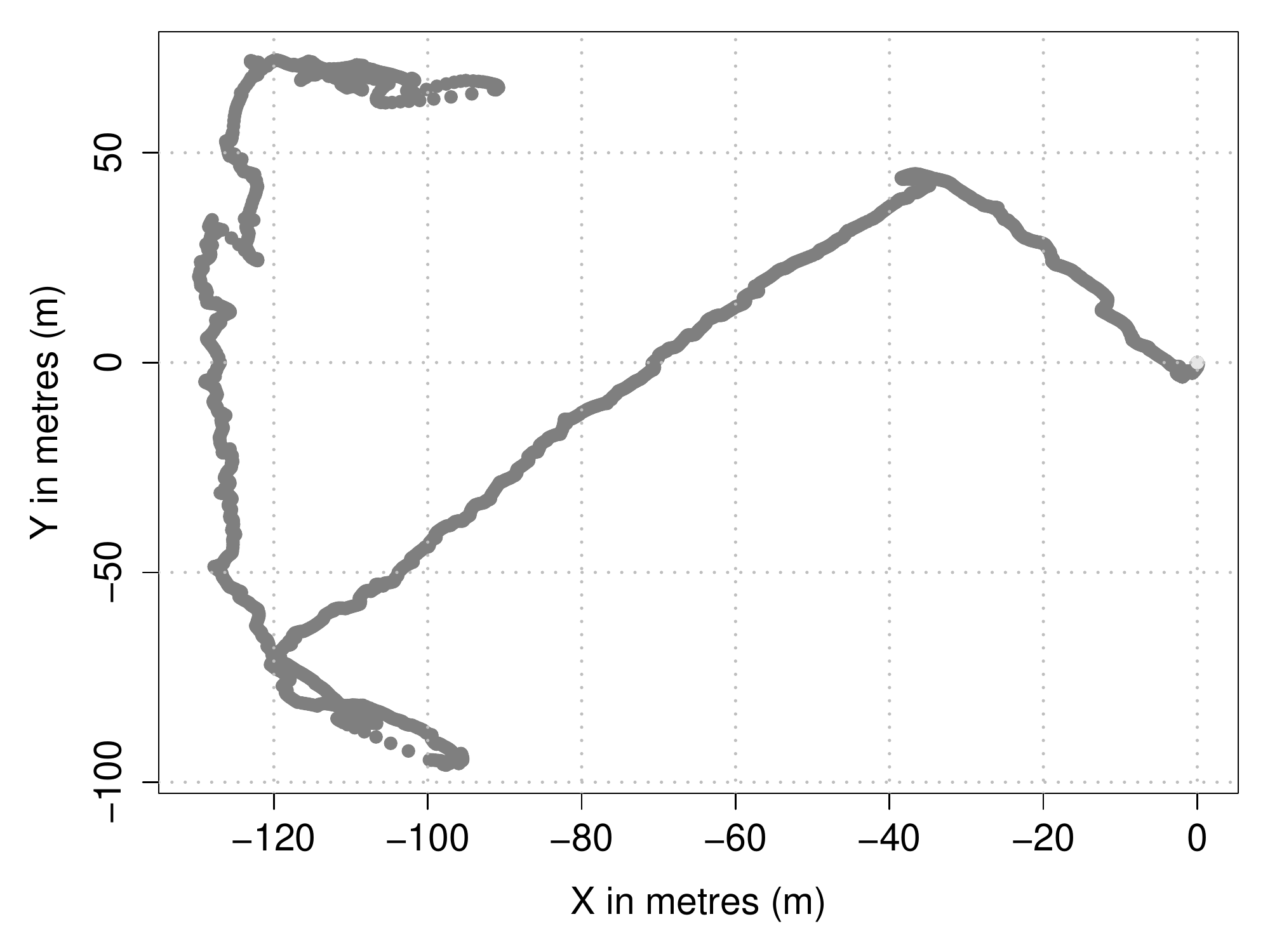}}
    \subfigure[]{\includegraphics[width=\2]{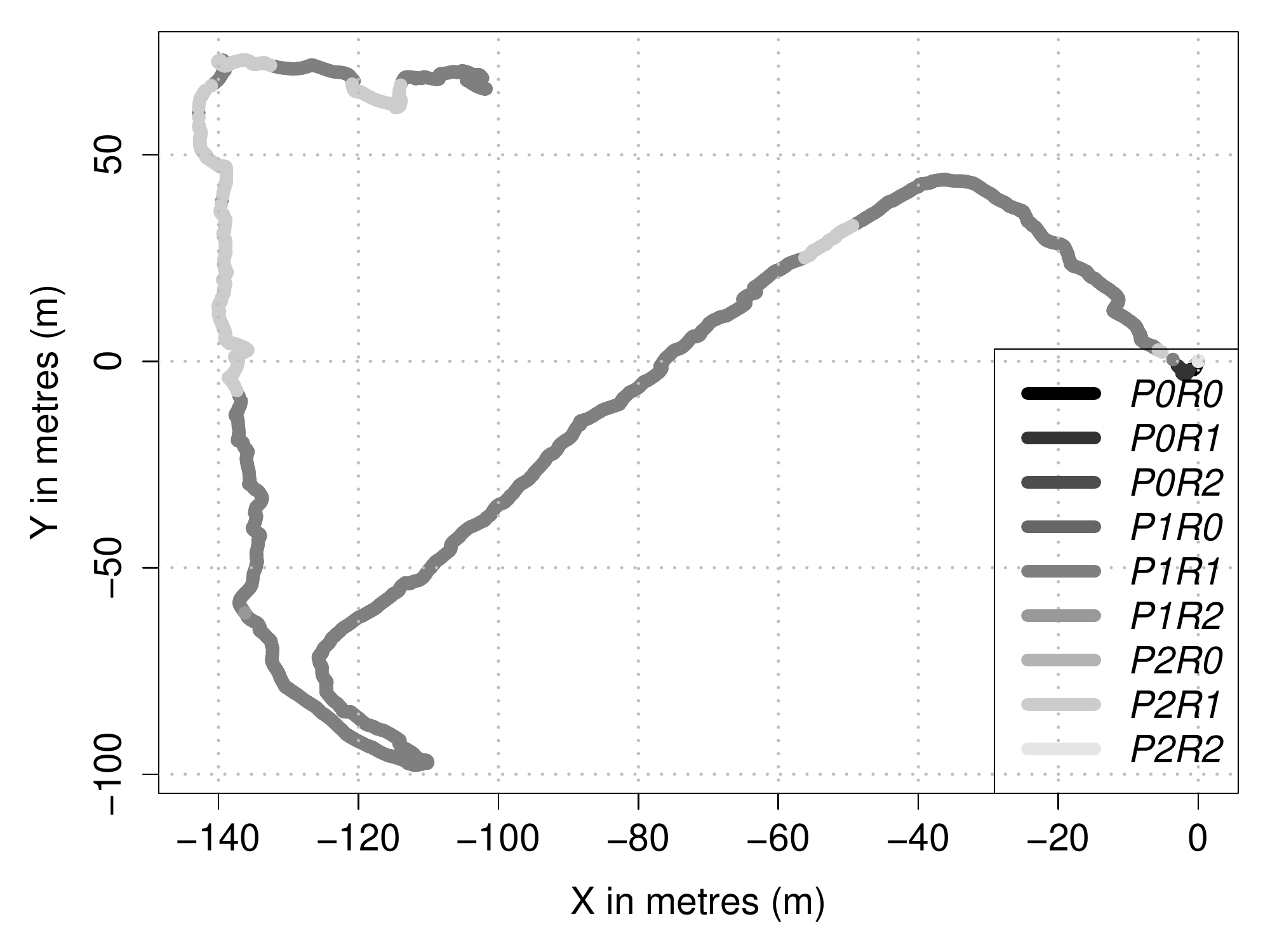}}
    \subfigure[]{\includegraphics[width=\2]{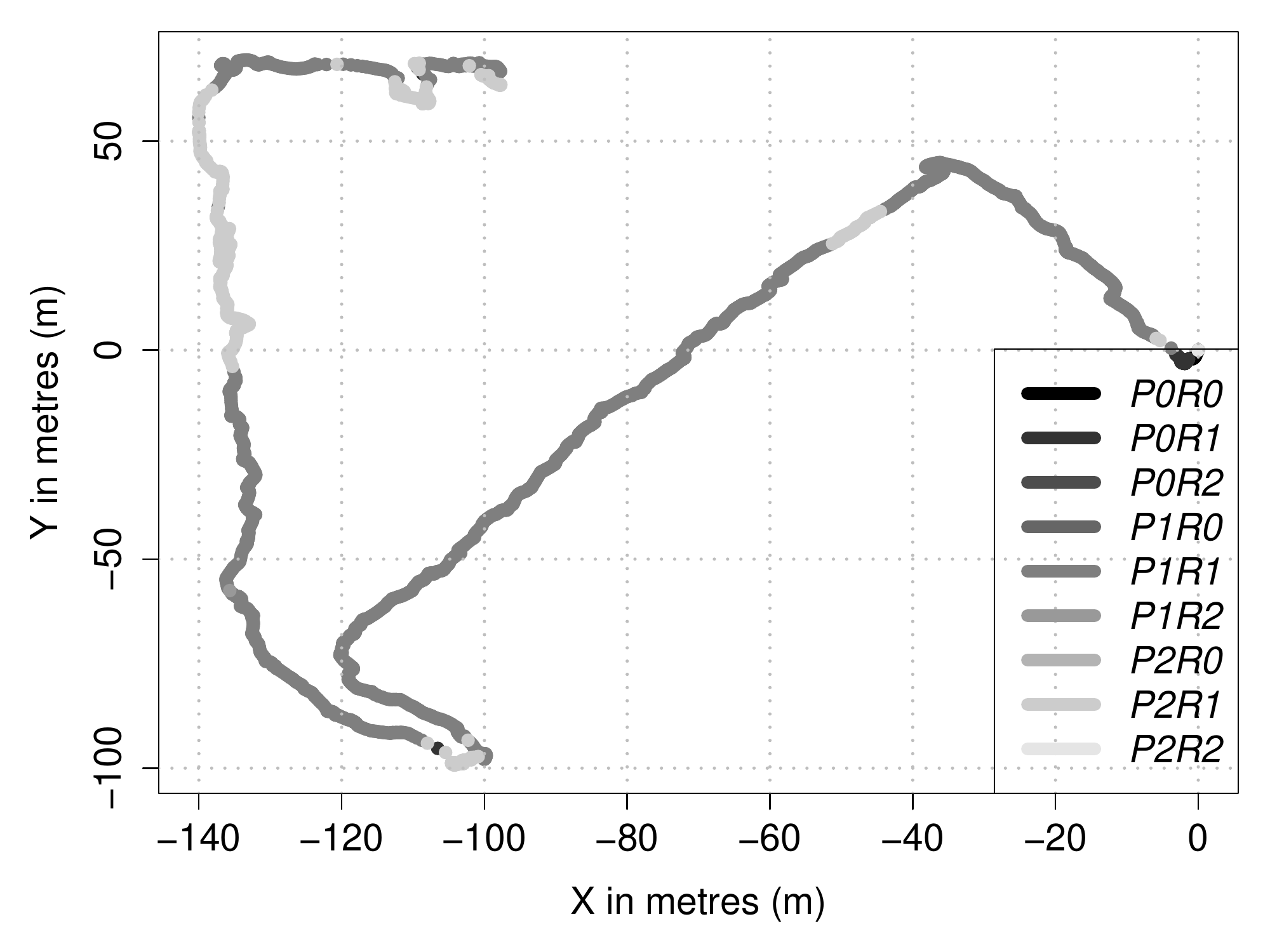}}
  \end{center}
  \caption[Real and estimated paths for dataset 1]{Real and estimated paths for dataset 1. Colours indicate the type of the motion model employed for estimating a part of the path. (a) Ground truth data (b) Path using GPS and IMU with CMM (c) Path using GPS, camera and IMU with CMM (d) Path using GPS and IMU with FAMM (e) Path using GPS, camera and IMU with FAMM}
  \label{fig:trajectoryDataset1}
\end{figure}

\begin{figure}
  \begin{center}
    \subfigure[]{\includegraphics[width=\2]{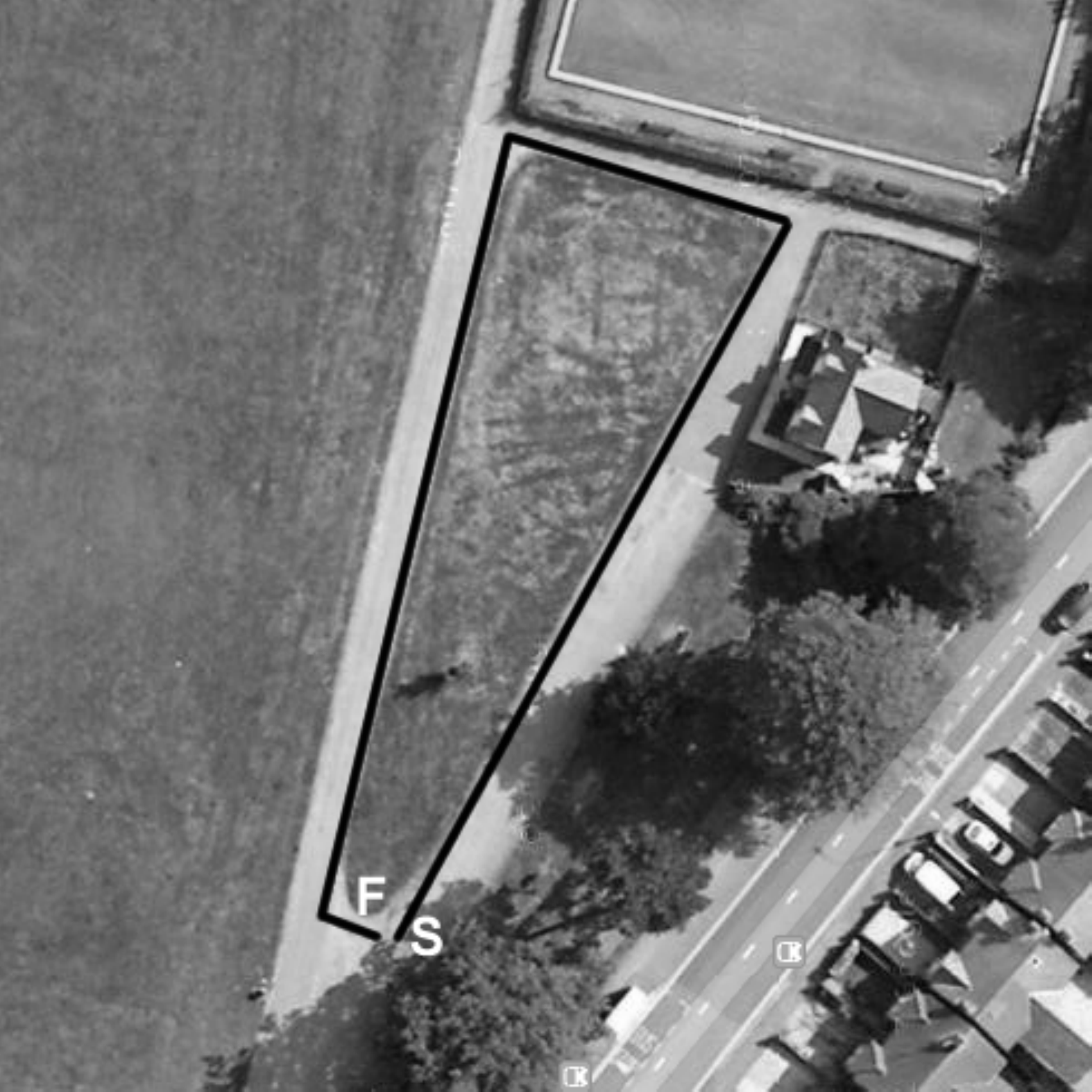}}\\
    \subfigure[]{\includegraphics[width=\2]{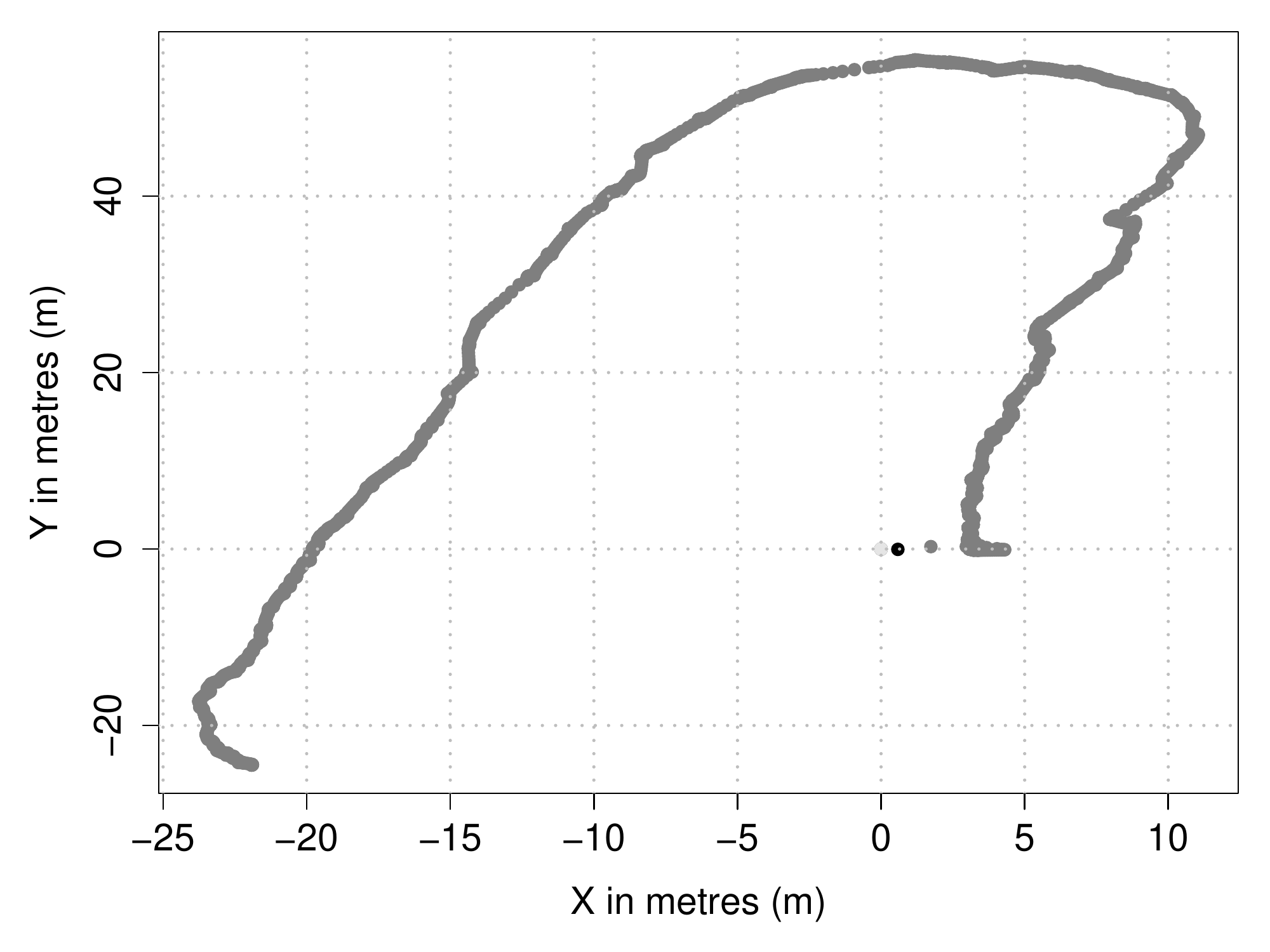}}
    \subfigure[]{\includegraphics[width=\2]{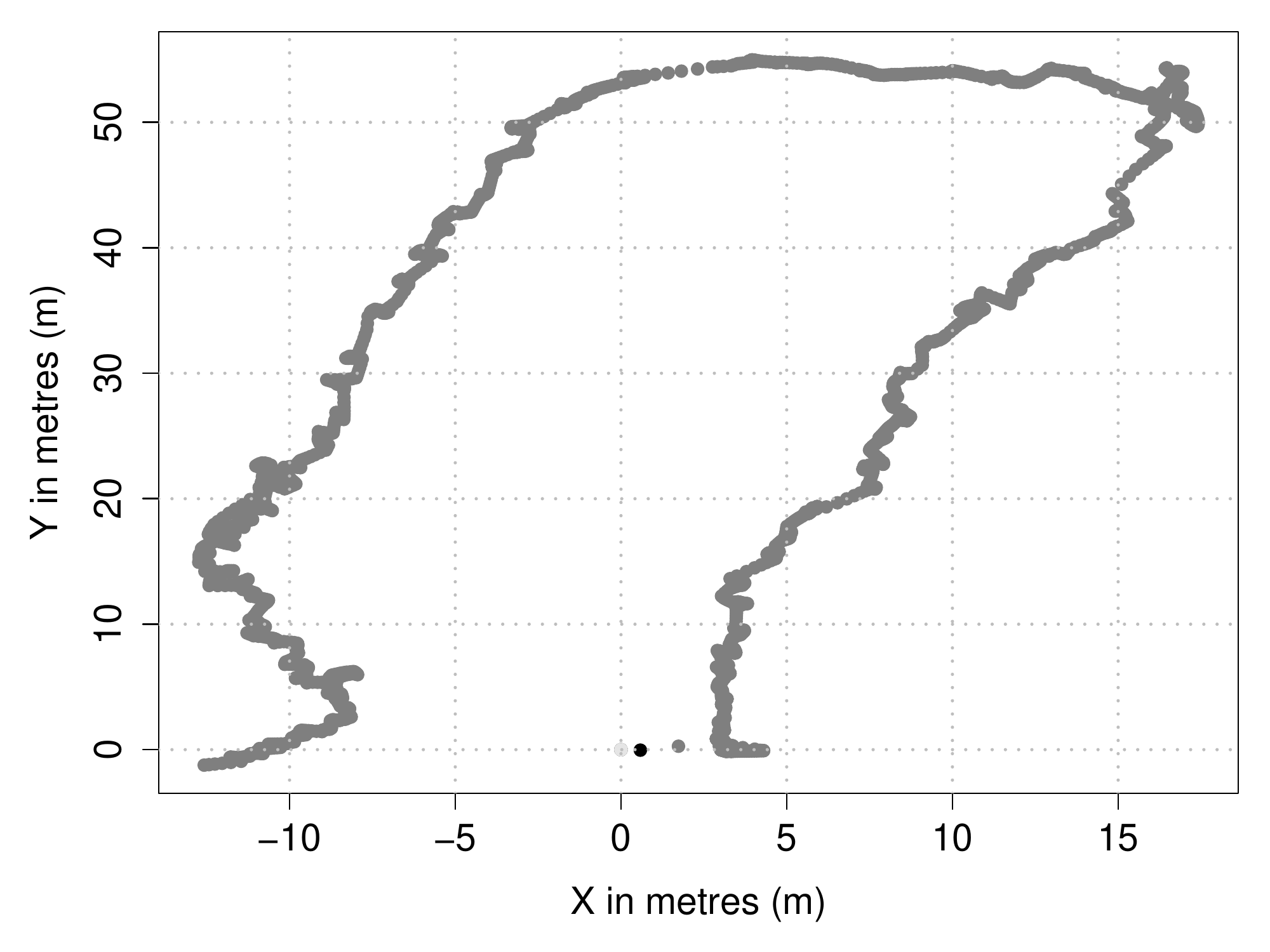}}
    \subfigure[]{\includegraphics[width=\2]{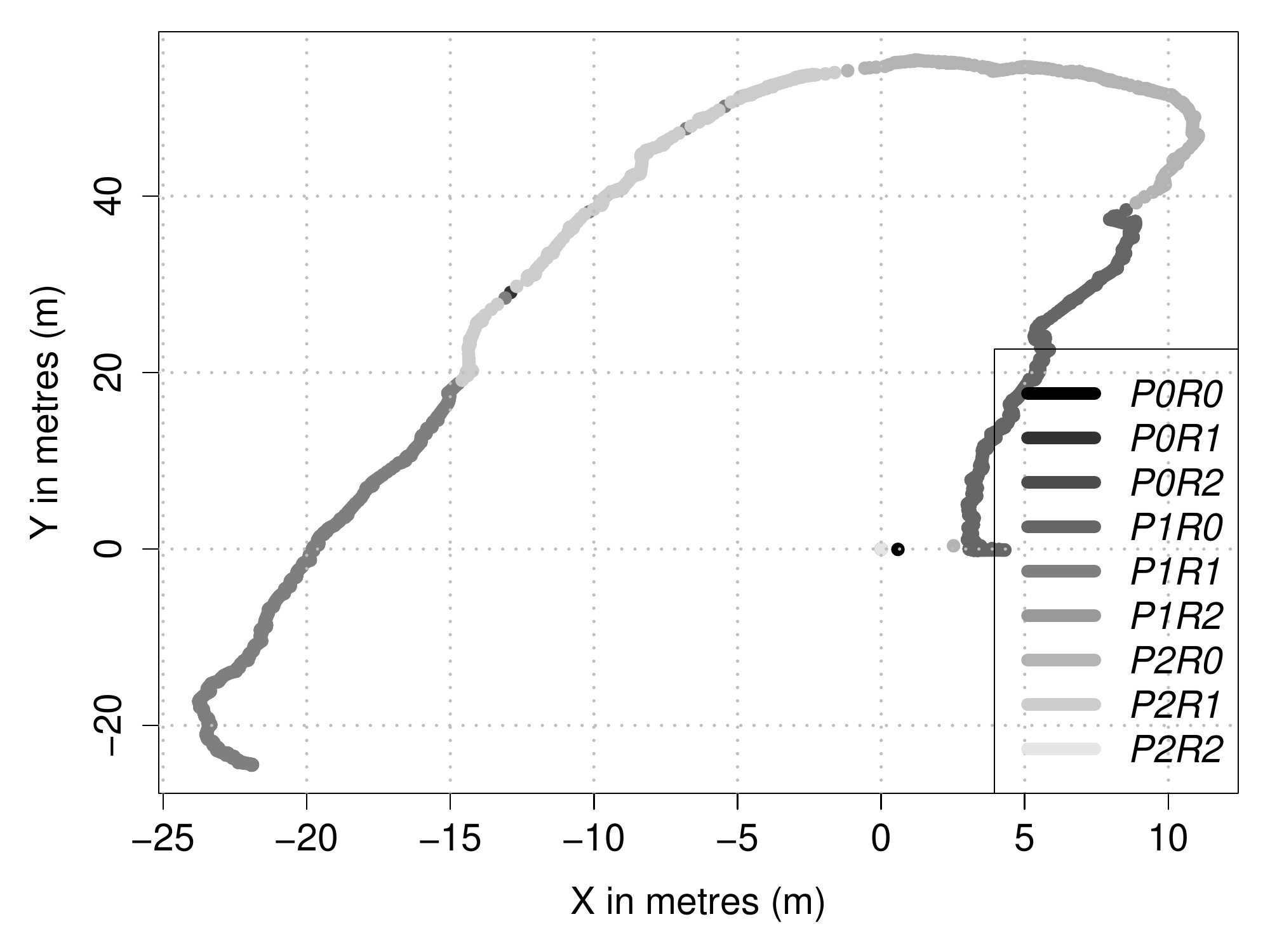}}
    \subfigure[]{\includegraphics[width=\2]{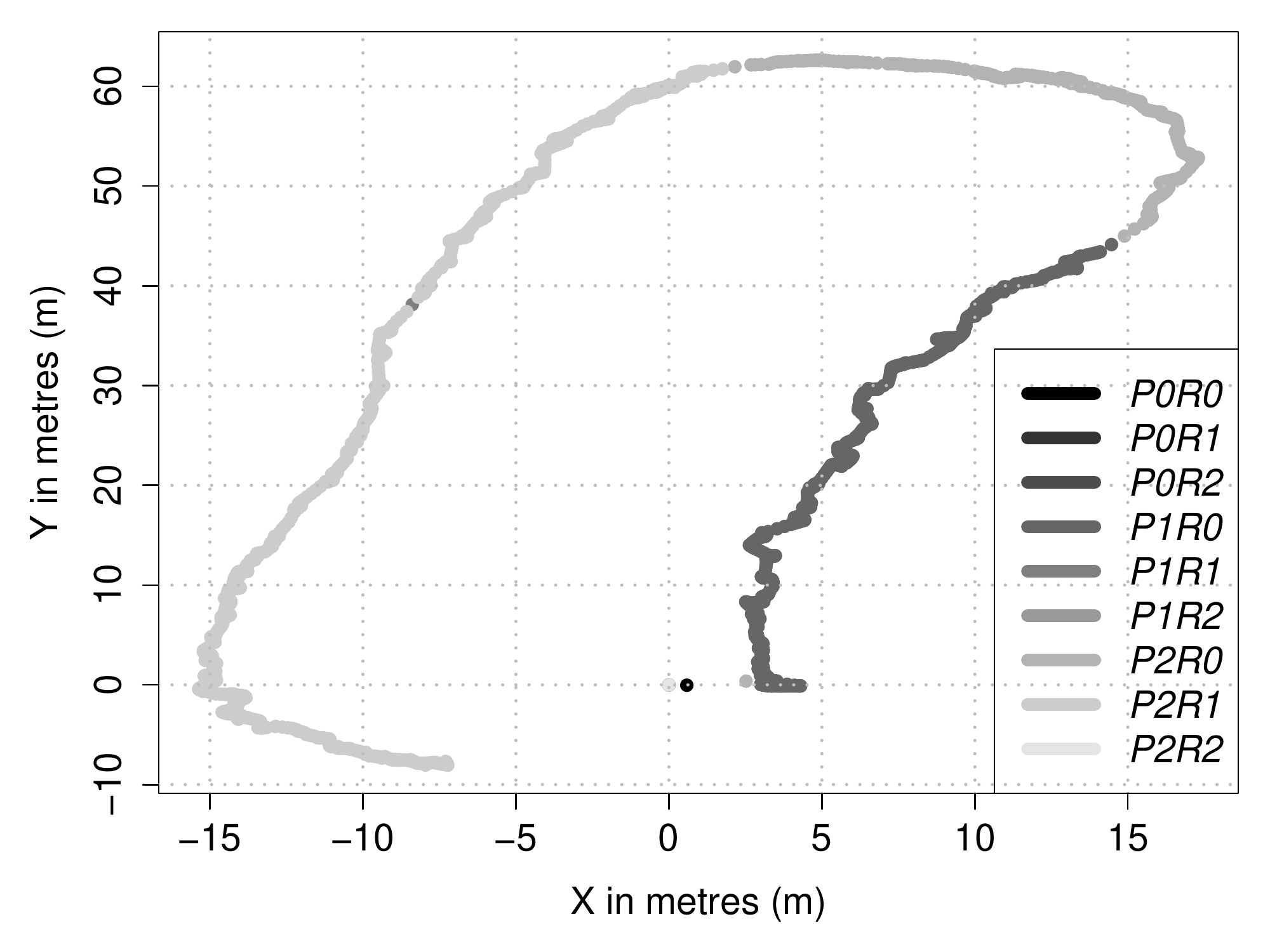}}
  \end{center}
  \caption[Real and estimated paths for dataset 5]{Real and estimated paths for dataset 5. Colours indicate the type of the motion model employed for estimating a part of the path. (a) Ground truth data (b) Path using GPS and IMU with CMM (c) Path using GPS, camera and IMU with CMM (d) Path using GPS and IMU with FAMM (e) Path using GPS, camera and IMU with FAMM}
    \label{fig:trajectoryDataset5}
\end{figure}

Figures~\ref{fig:trajectoryDataset1} to~\ref{fig:trajectoryDataset5} show the ground-truth path and the estimated paths using integration of different sensors and employing different motion models based on the FAMM presented in Section~\ref{sec:fuzzyMotionModels}. Portions of the estimated paths are coloured differently, emphasizing the type of the motion model used for estimation. It is important to note that the CMM used in the figures correspond to \texttt{P1R1} and hence is drawn in the same colour.

The advantage of using a camera and the FAMM is clearly shown in Figure~\ref{fig:trajectoryDataset5}, an example case for loop--closing. The integration of GPS and IMU could not handle the last segment of the path both when CMM (b) and FAMM (d) are used. In (c), the last segment was identified using integration of the camera with the two other sensors; however, the direction was not correct. Employing FAMM with the three sensors, shown in (e), gives the most accurate estimation. Furthermore, the overall shape of the estimated trajectory is closest to the ground-truth path.

\begin{figure}
  \begin{center}   
    \subfigure[]{\includegraphics[width=\1]{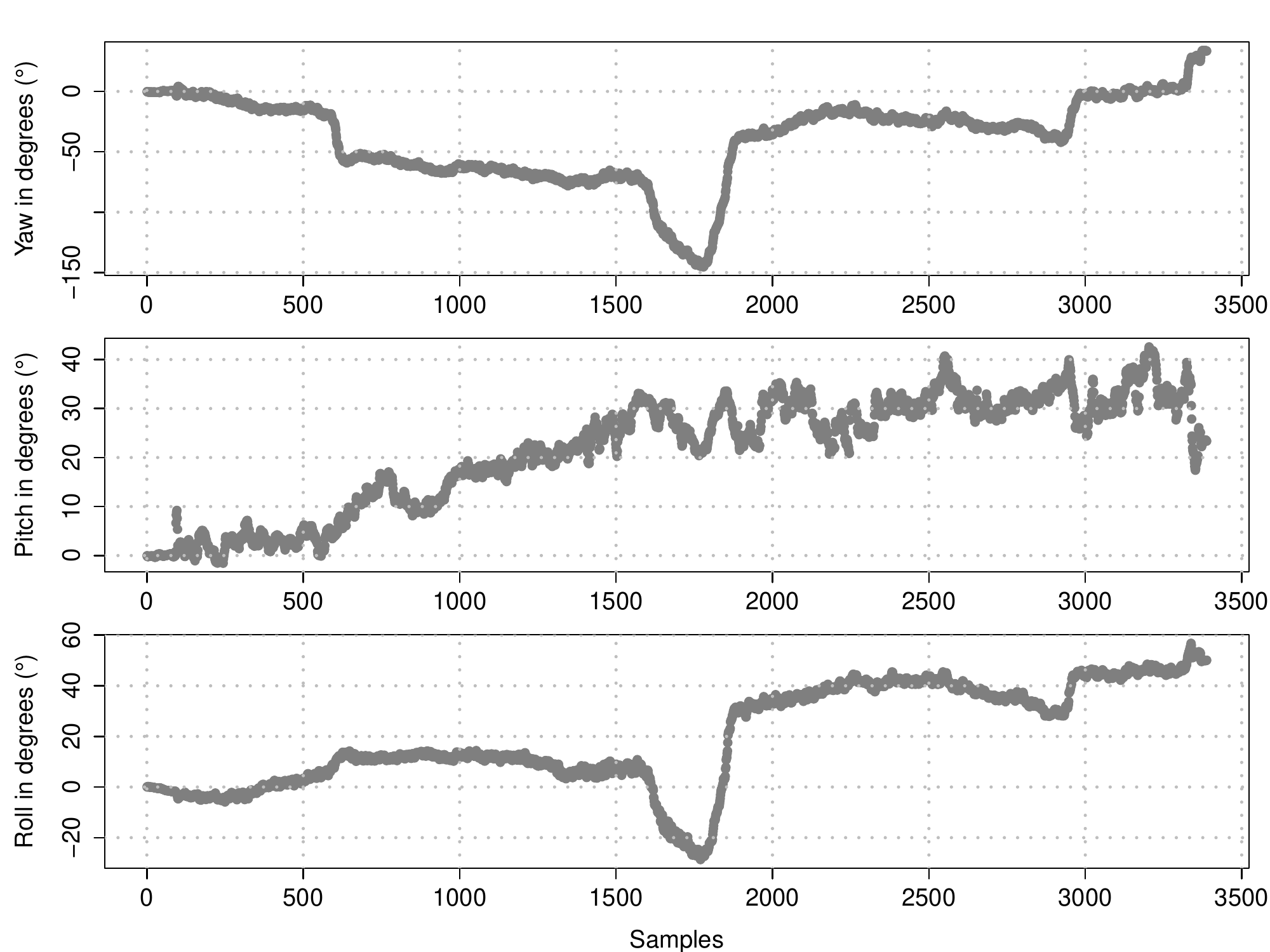}}\\
    \subfigure[]{\includegraphics[width=\1]{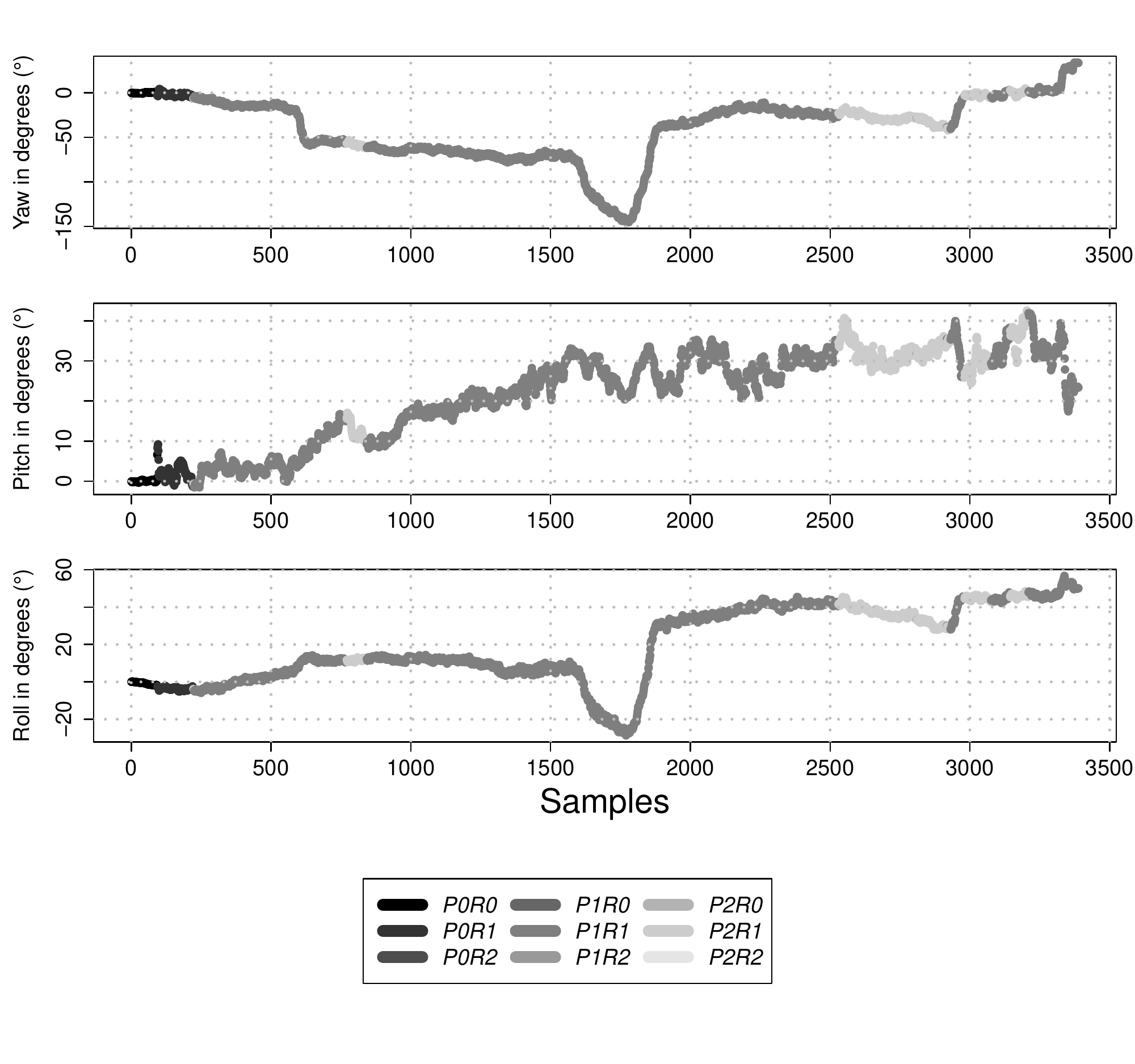}}
  \end{center}
  \caption[Estimated rotations for CMM and FAMM for dataset 1]{Estimated rotations for CMM and FAMM for dataset 1. Colours indicate the type of the motion model used to estimate the orientation. (a) Orientation using CMM (b) Orientation using FAMM}
  \label{fig:RotationsDataset1}
\end{figure}

\begin{figure}
  \begin{center}   
    \subfigure[]{\includegraphics[width=\1]{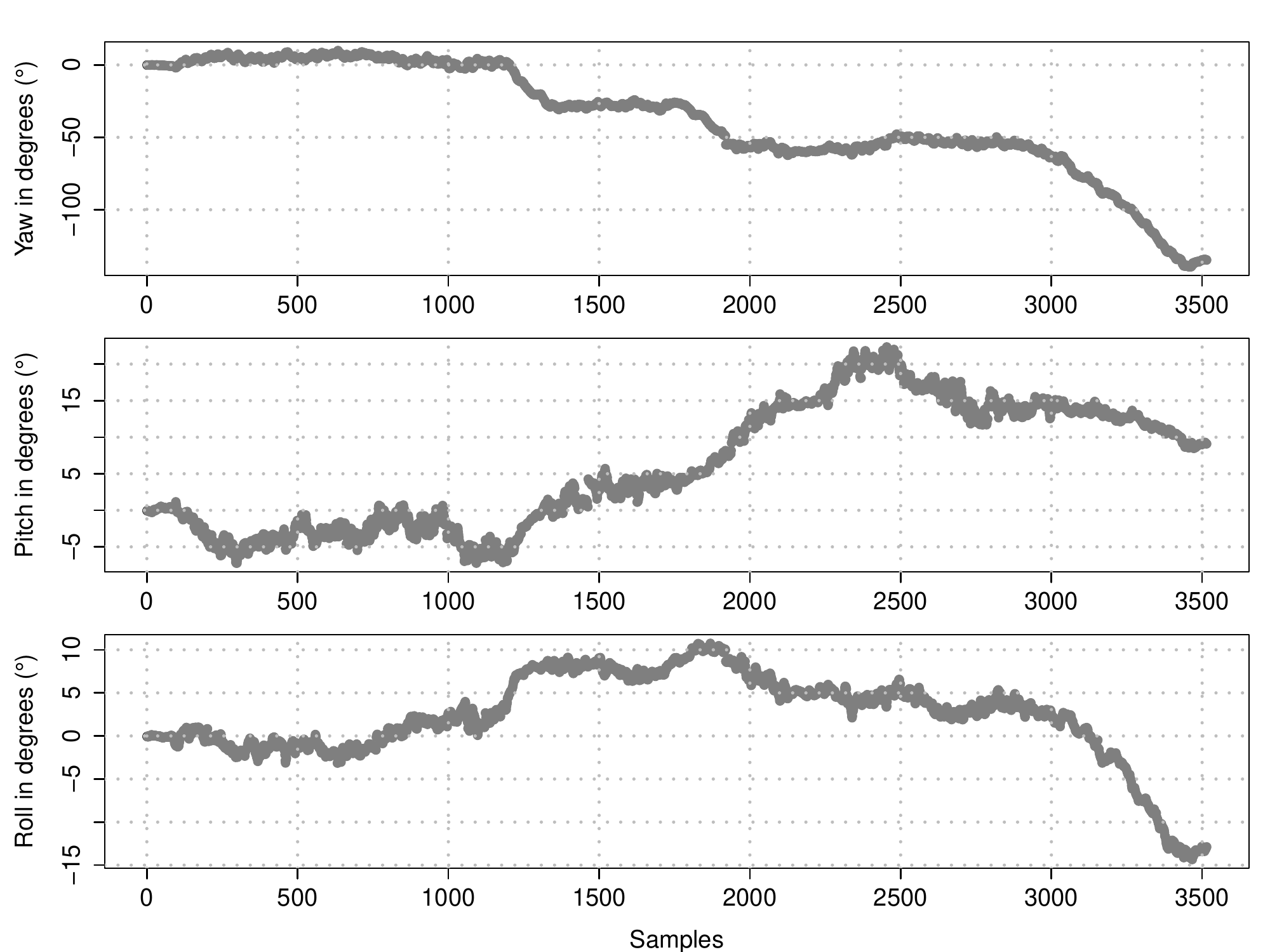}}\\
    \subfigure[]{\includegraphics[width=\1]{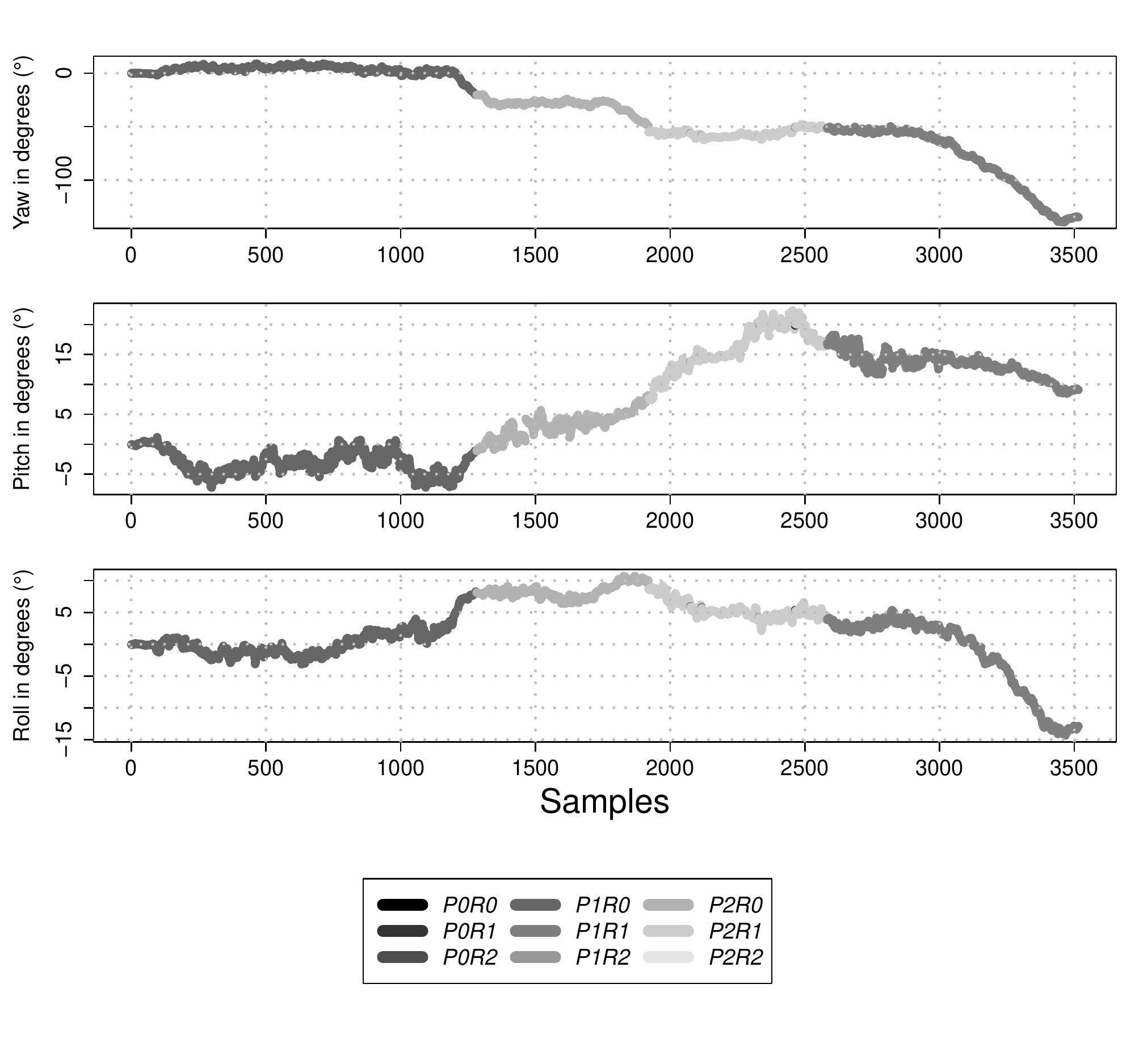}}
  \end{center}
  \caption[Estimated rotations for CMM and FAMM for dataset 5]{Estimated rotations for CMM and FAMM for dataset 5. Colours indicate the type of the motion model used to estimate the orientation. (a) Orientation using CMM (b) Orientation using FAMM}
    \label{fig:RotationsDataset5}
\end{figure}

Figures~\ref{fig:RotationsDataset1} to~\ref{fig:RotationsDataset5} present the estimated orientations for the datasets using the CMM and FAMM. One thing to mention in these orientation plots is that there is less jitter when the FAMM is employed for the motion model.

The aim of using multiple motion models was to decrease the uncertainty in the filter due to user motion, which is not always predictable. This observed decrease is mainly due to the selection of the most appropriate motion model, better fitting the measurements providing more supporting evidence for the filter so that is more certain of its current state. The state covariance matrix of a Kalman filter ($\Sigma$)  includes this estimate of uncertainty as well as correlations between state vector ($x$) elements. The diagonal elements indicate the variances and off-diagonal ones store correlations~\cite{Kleeman2013}. This matrix uses the information provided to the filter through the Kalman gain ($K$) indirectly from the measurements. 

\begin{figure}
  \begin{center}   
    \subfigure[]
    {\includegraphics[width=\2]{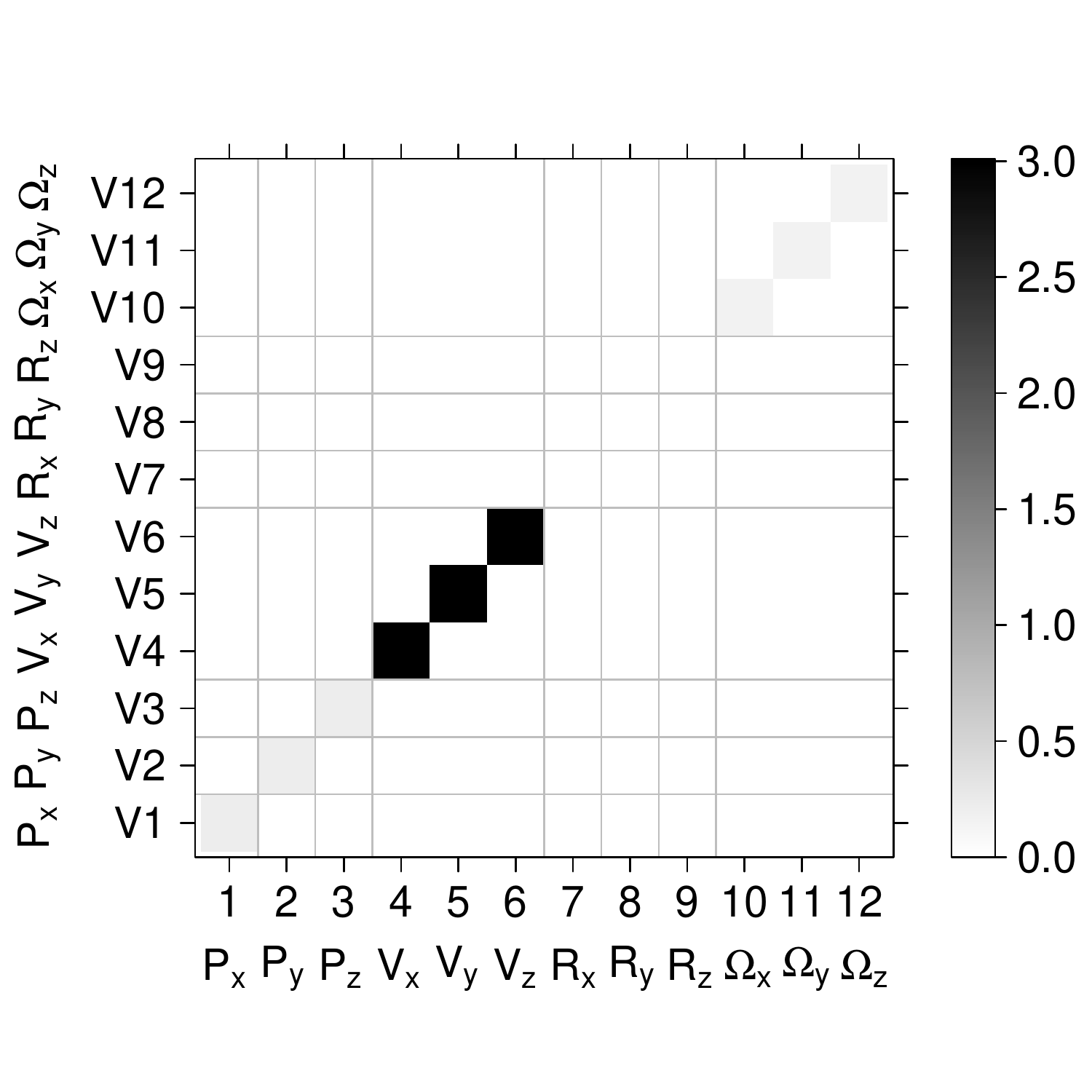}}
    \subfigure[]
    {\includegraphics[width=\2]{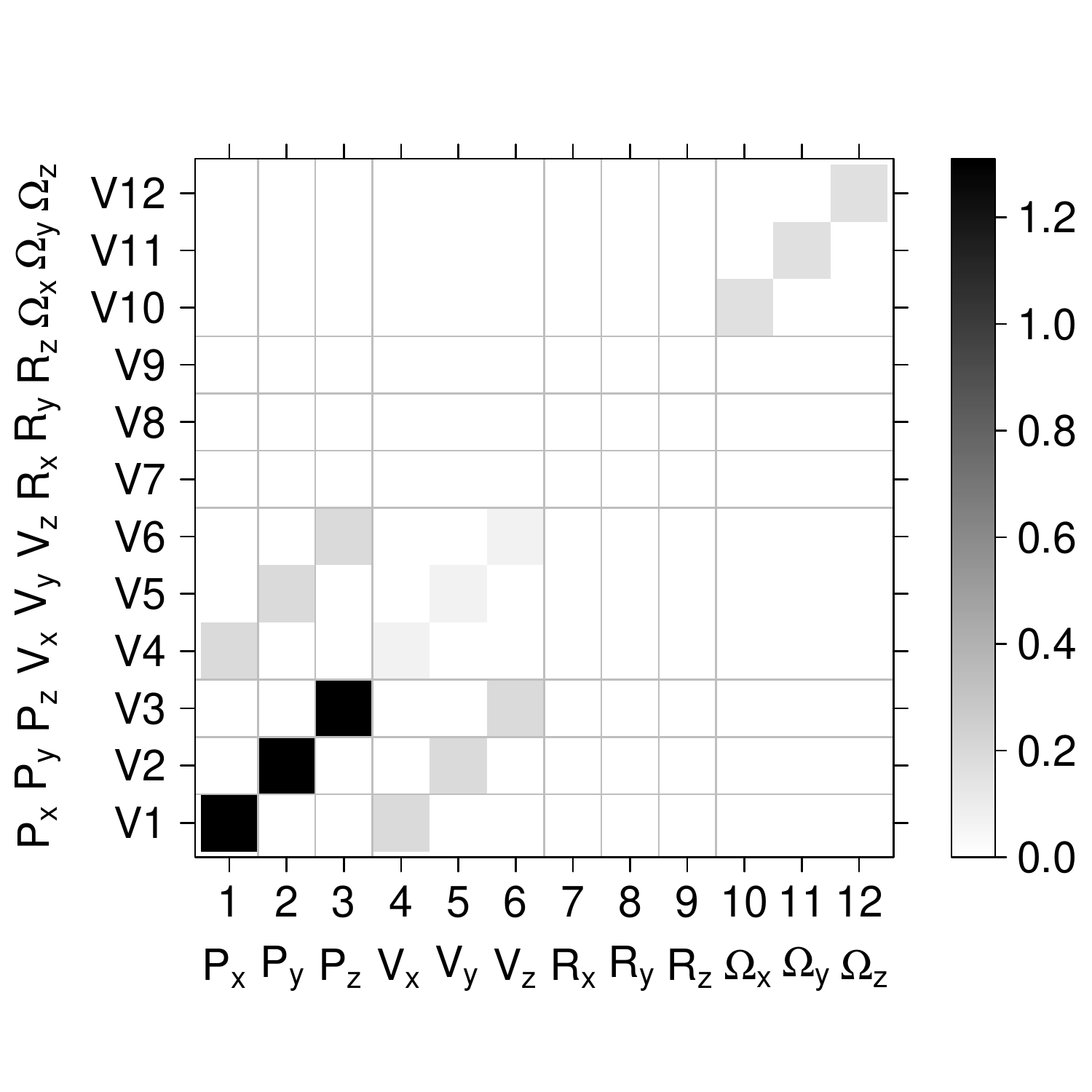}}    
    \subfigure[]
    {\includegraphics[width=\2]{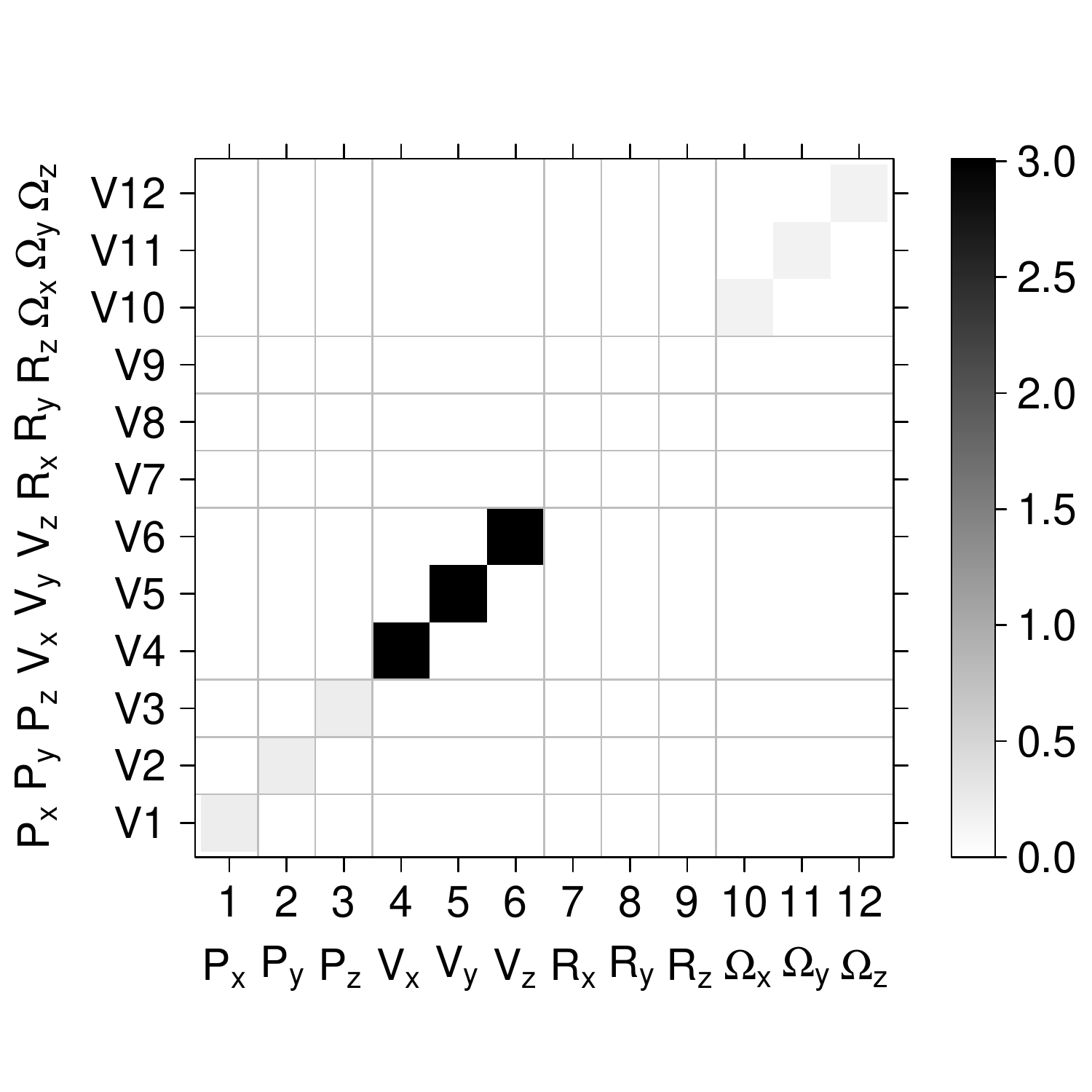}}
    \subfigure[]
    {\includegraphics[width=\2]{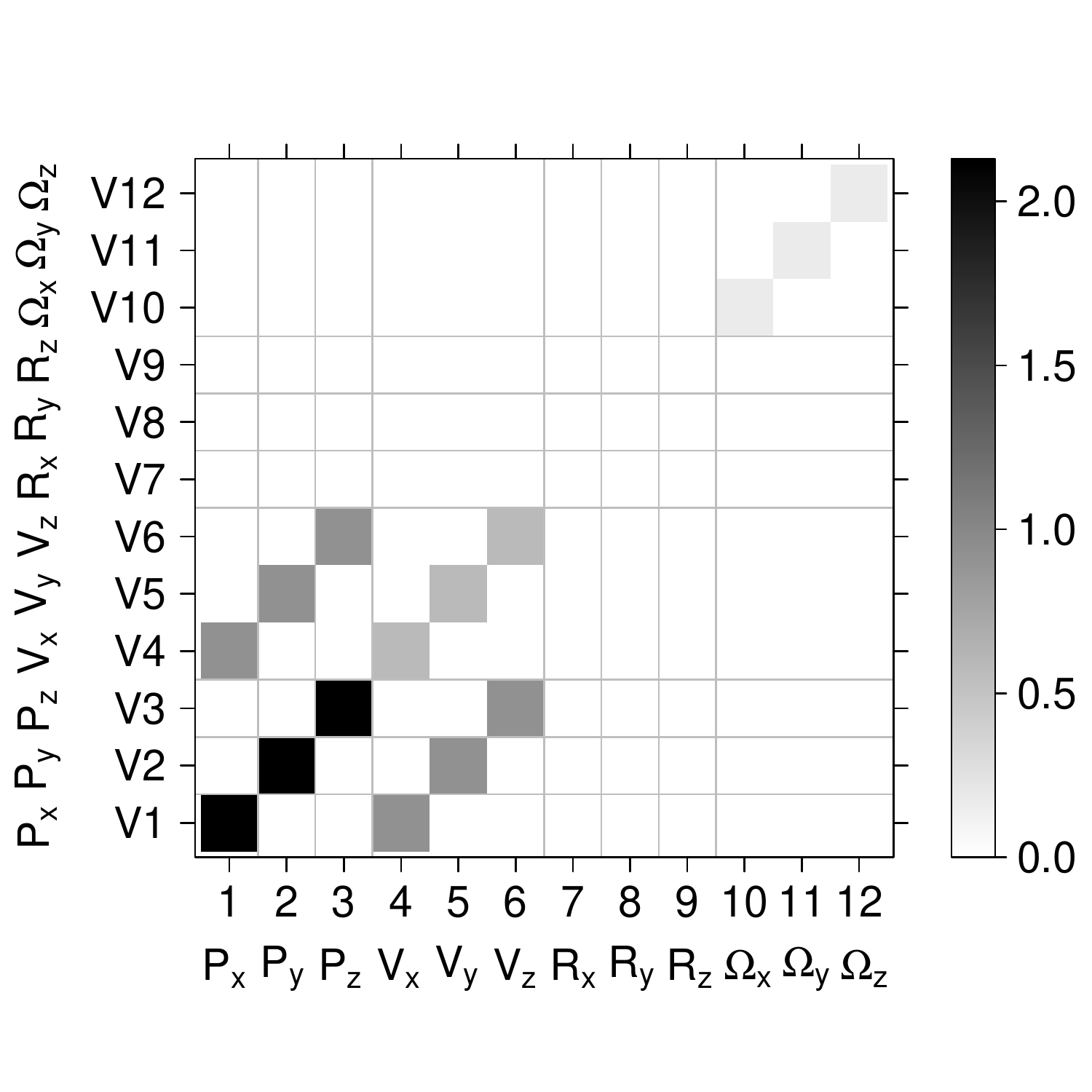}}      \end{center}
  \caption[Changes in the state covariances for dataset 1]{Changes in the state covariances for dataset 1 when CMM and FAMM are employed for GPS--IMU and camera--GPS--IMU integration. Colours indicate the magnitude of the covariance matrix elements. Amount of uncertainty is illustrated by higher magnitudes (darker colours). (a) GPS--IMU with CMM (b) GPS--IMU with FAMM (c) Camera--GPS--IMU with CMM (d) Camera--GPS--IMU with FAMM}
  \label{fig:stateCovDataset1}
\end{figure}

\begin{figure}
  \begin{center}   
    \subfigure[]
    {\includegraphics[width=\2]{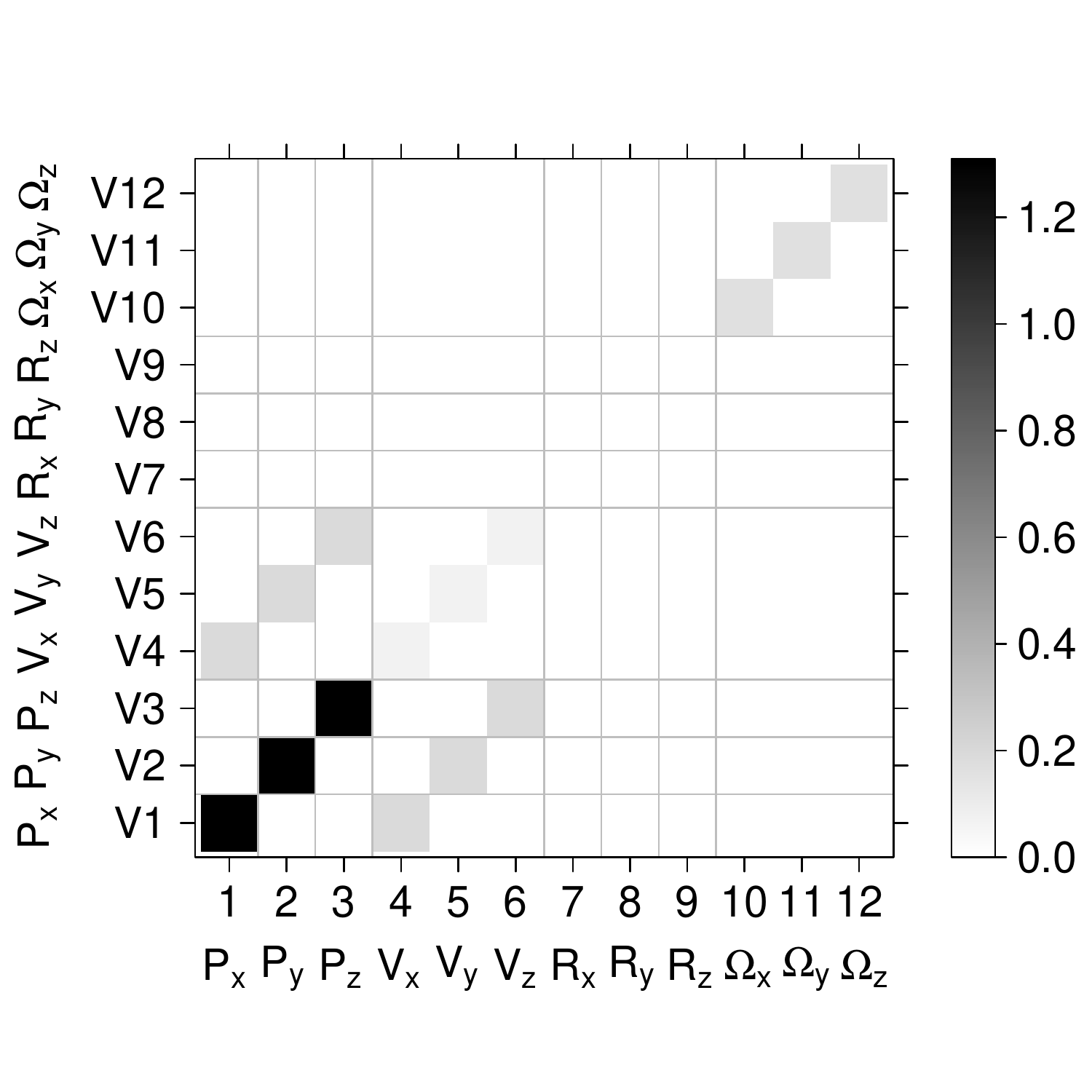}}
    \subfigure[]
    {\includegraphics[width=\2]{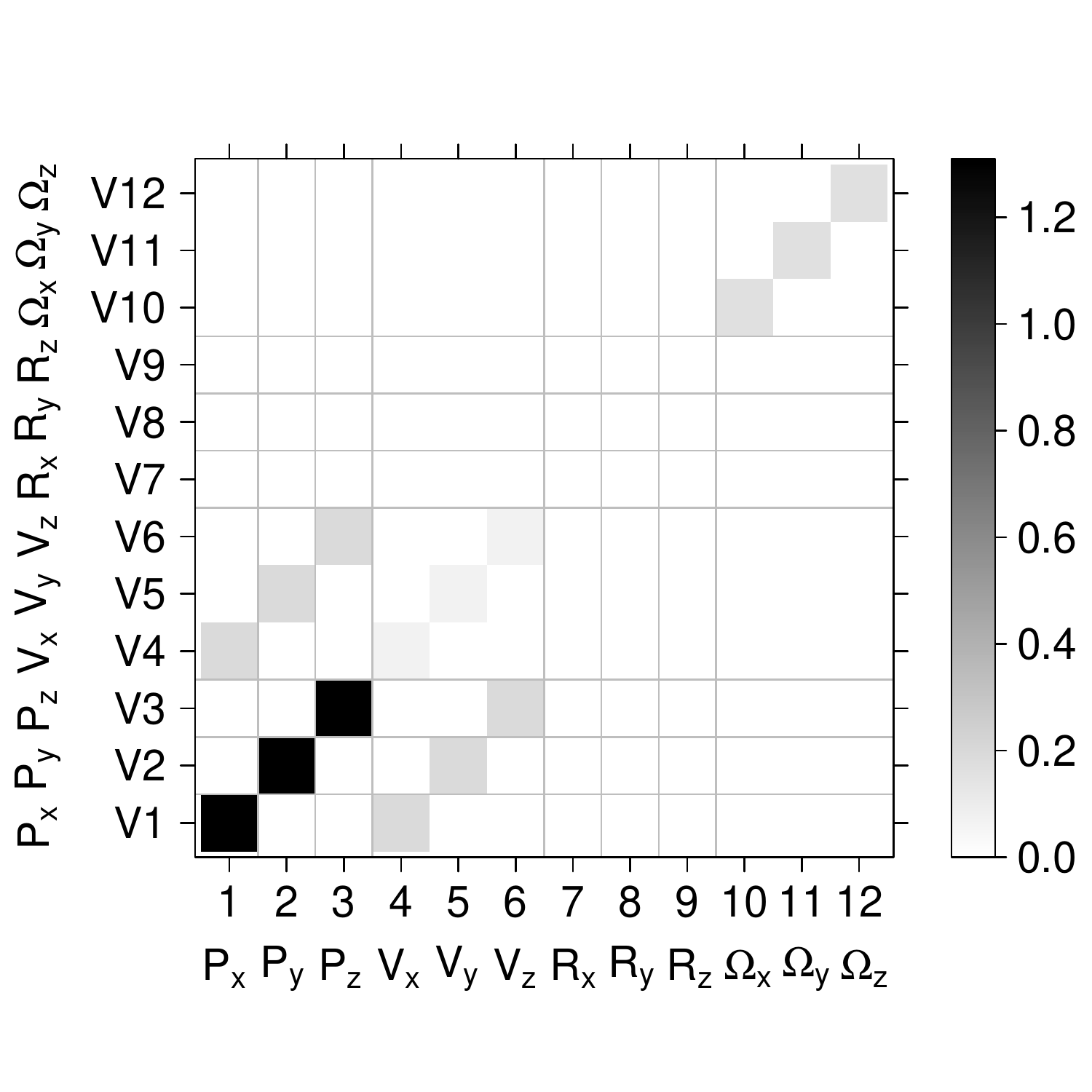}}    
    \subfigure[]
    {\includegraphics[width=\2]{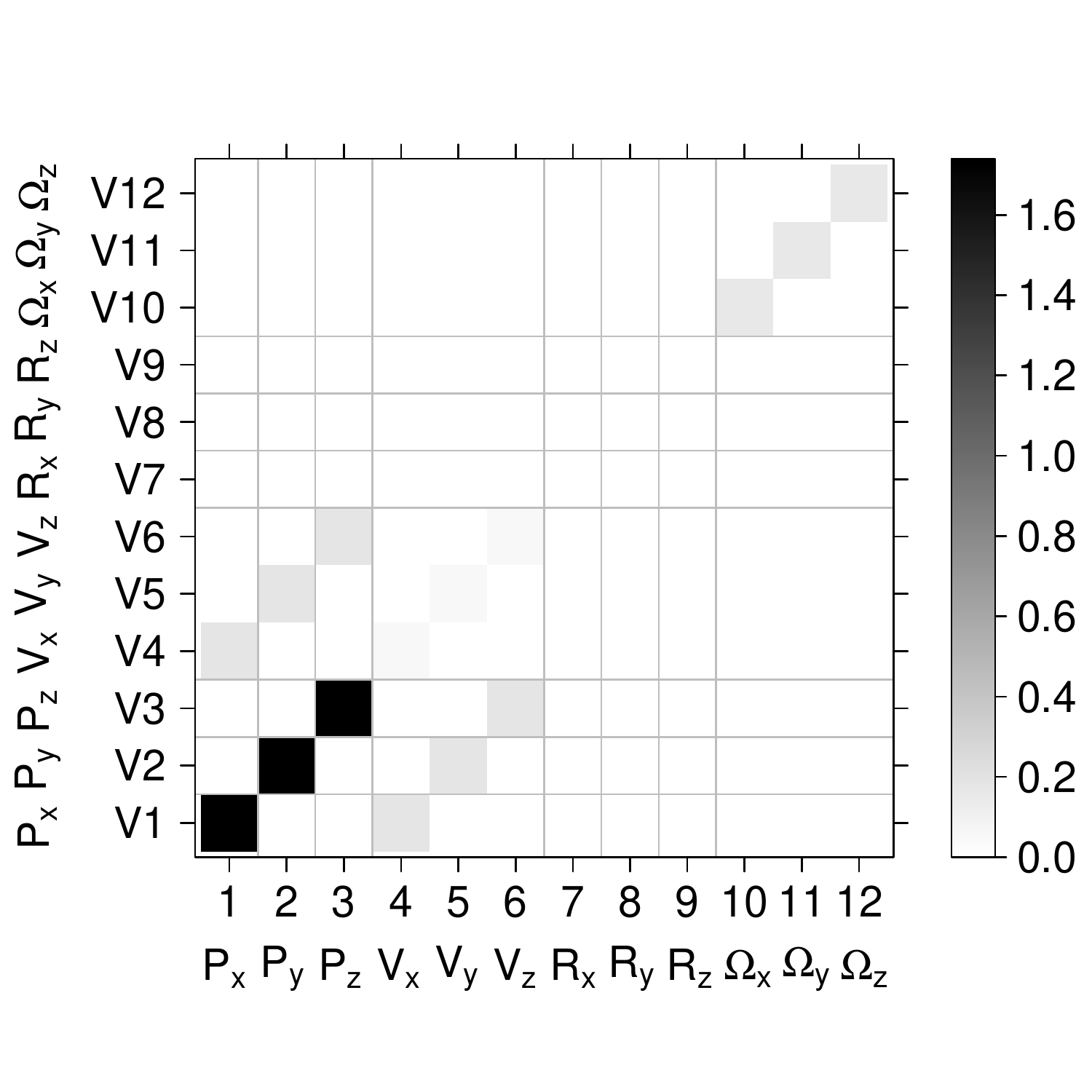}}
    \subfigure[]
    {\includegraphics[width=\2]{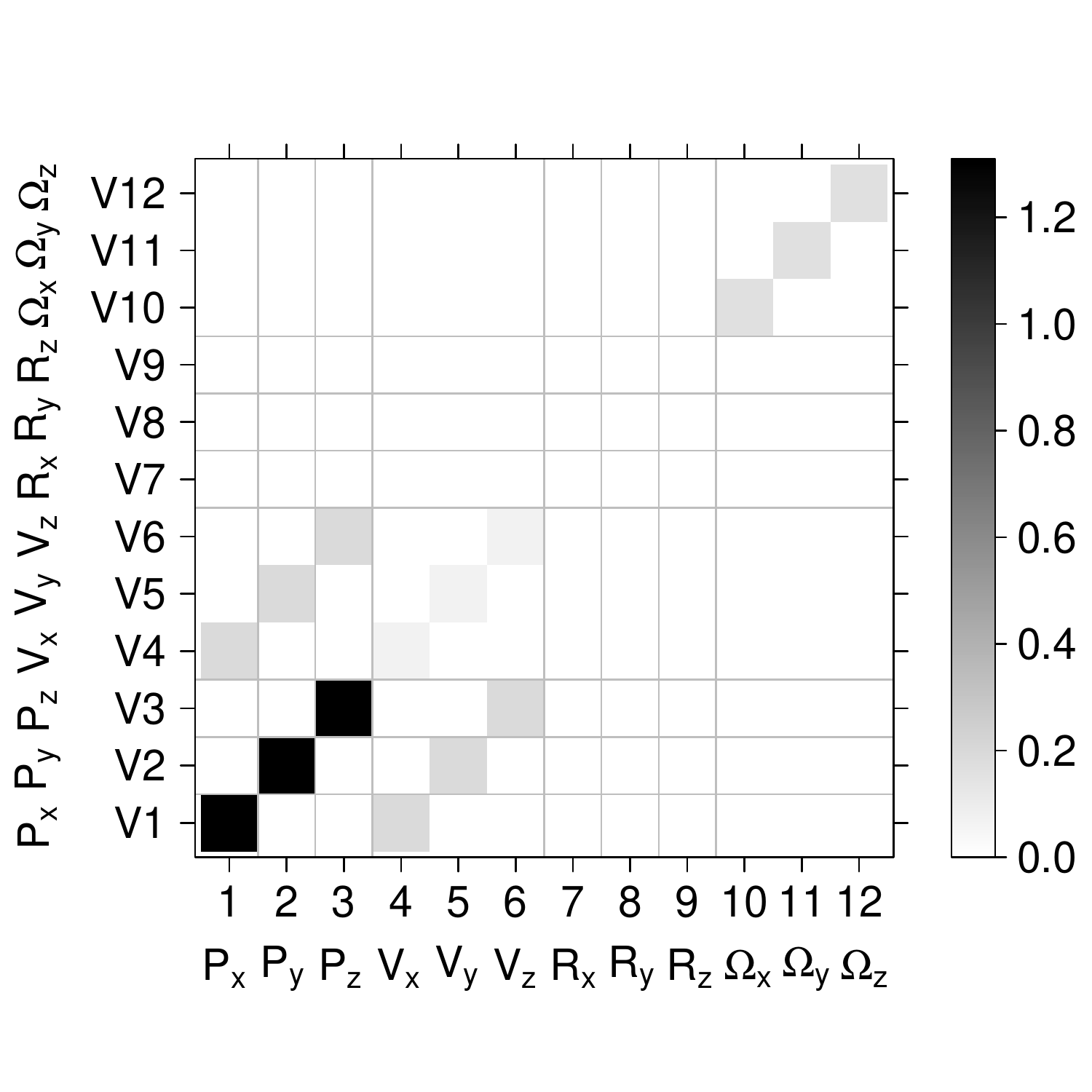}}      \end{center}
  \caption[Changes in the state covariances for dataset 5]{Changes in the state covariances for dataset 5 when CMM and FAMM are employed for GPS--IMU and camera--GPS--IMU integration. Colours indicate the magnitude of the covariance matrix elements. Amount of uncertainty is illustrated by higher magnitudes (darker colours). (a) GPS--IMU with CMM (b) GPS--IMU with FAMM (c) Camera--GPS--IMU with CMM (d) Camera--GPS--IMU with FAMM}
  \label{fig:stateCovDataset5}
\end{figure}

The changes in the state covariance matrix are shown in Figures~\ref{fig:stateCovDataset1} to~\ref{fig:stateCovDataset5} where colours of the squares are associated with the magnitude of the matrix elements. The scales for colours in the diagonal elements for the state covariance matrix in the figures for FAMM indicate a decrease in system uncertainty when it is used instead of CMM.

\begin{figure}
  \begin{center}   
    \subfigure[]{\includegraphics[width=\1]{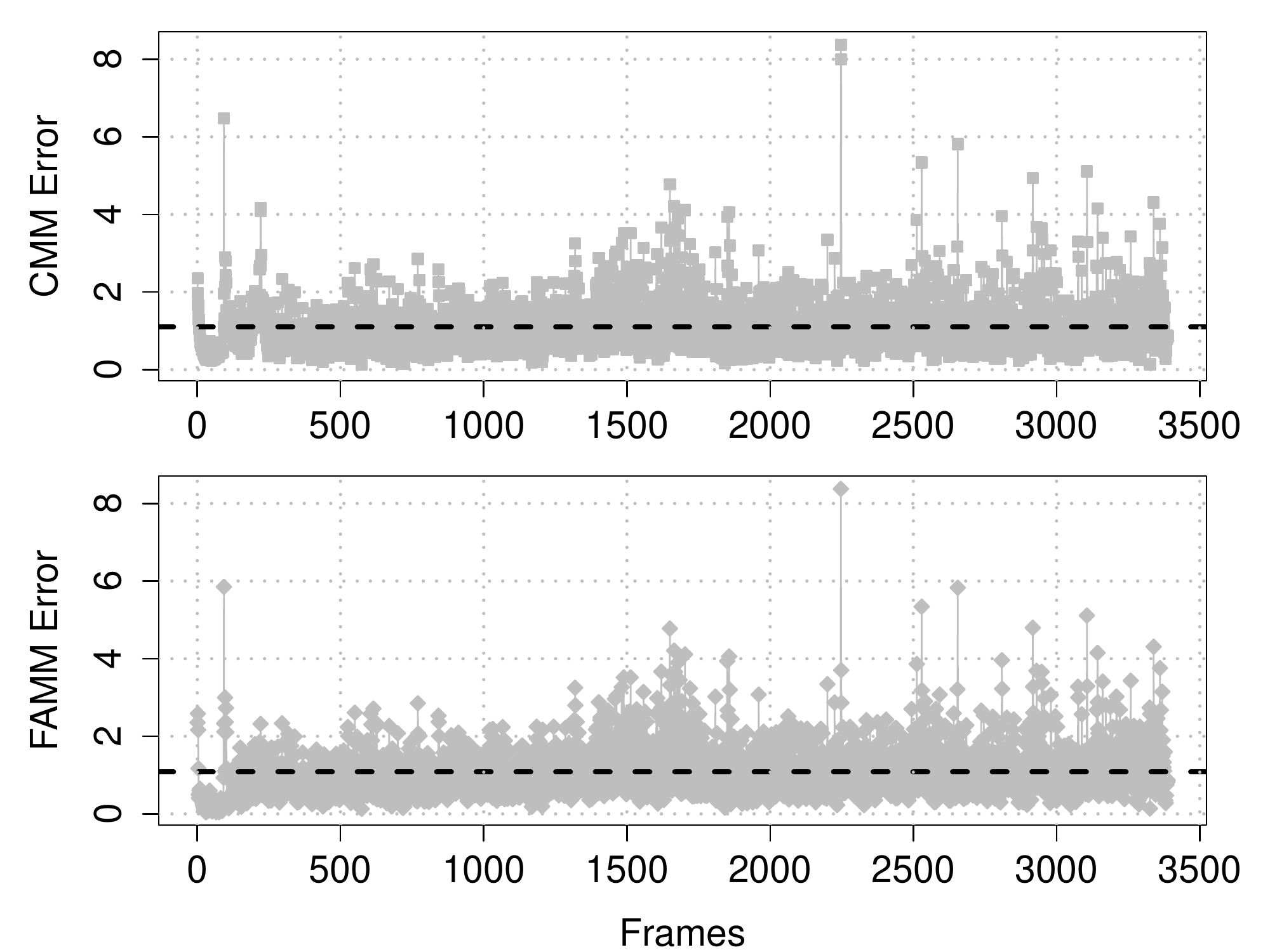}}
    \subfigure[]{\includegraphics[width=\1]{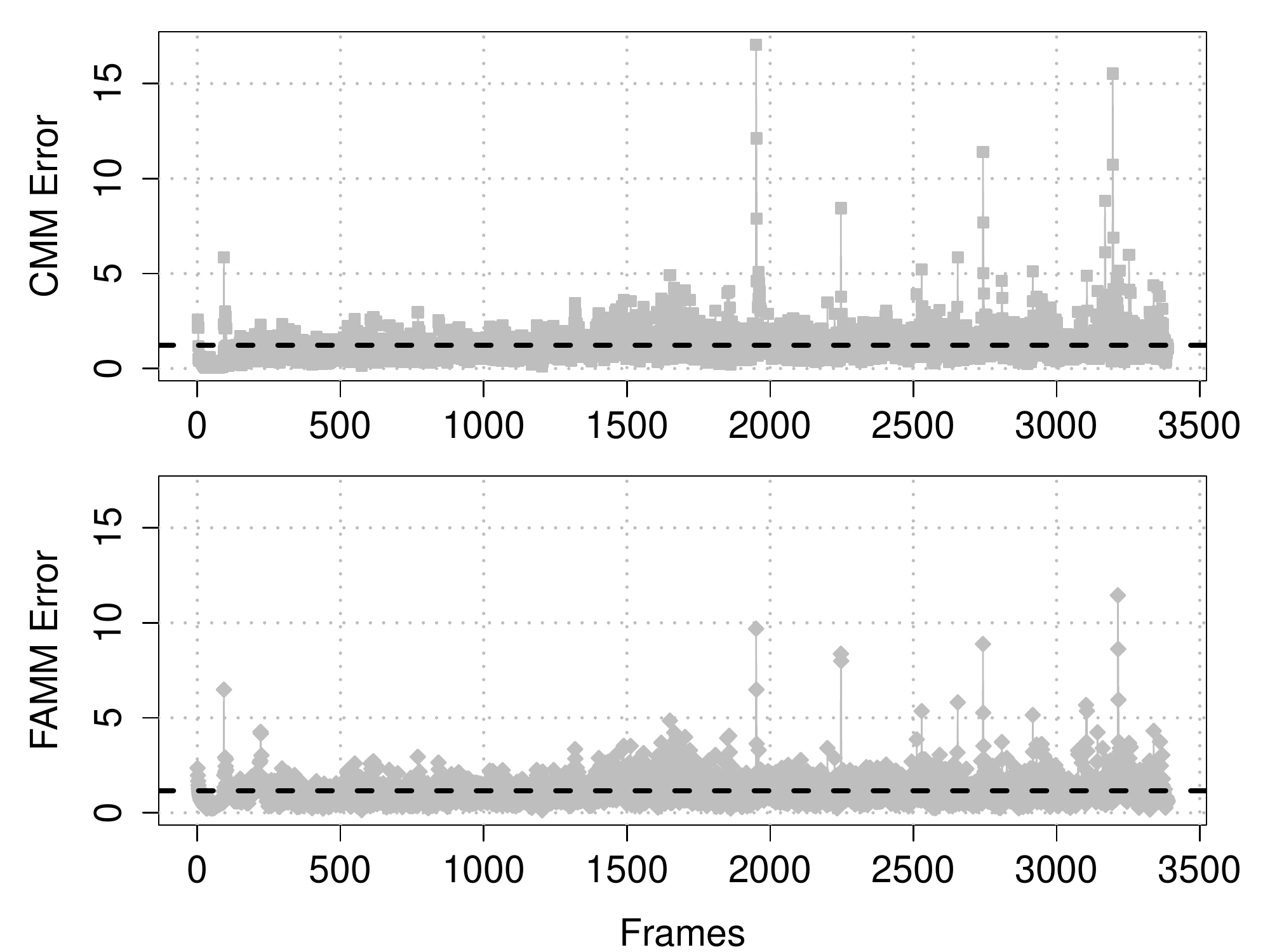}}
  \end{center}
  \caption[Filter errors for dataset 1 for CMM and FAMM]{Filter errors for dataset 1 for CMM and FAMM. The mean error is shown with the dashed line. (a) Filter error using GPS and IMU (b) Filter error using GPS, camera and IMU}
  \label{fig:filterErrorDataset1}
\end{figure}

\begin{figure}
  \begin{center}   
    \subfigure[]{\includegraphics[width=\1]{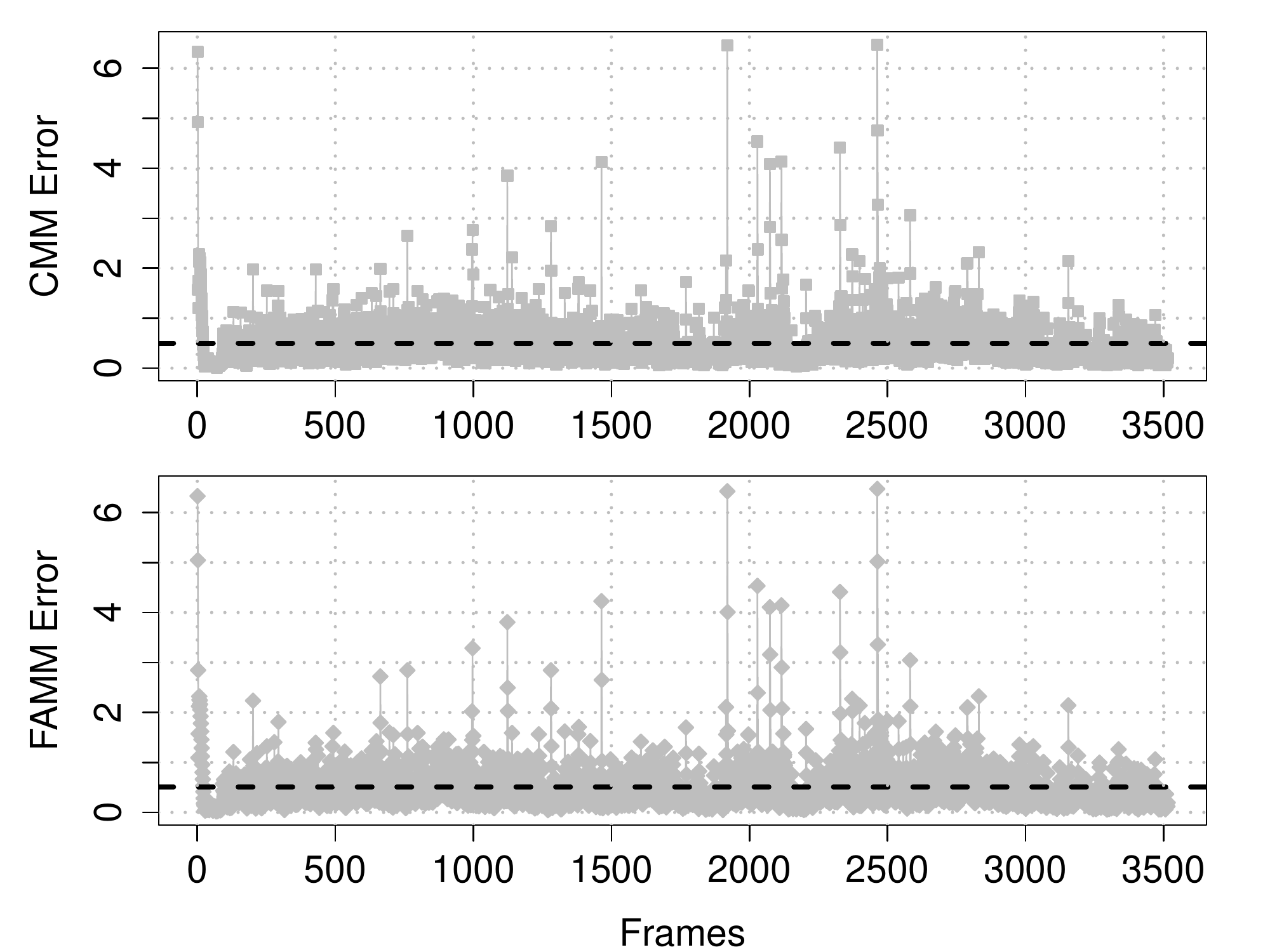}}
    \subfigure[]{\includegraphics[width=\1]{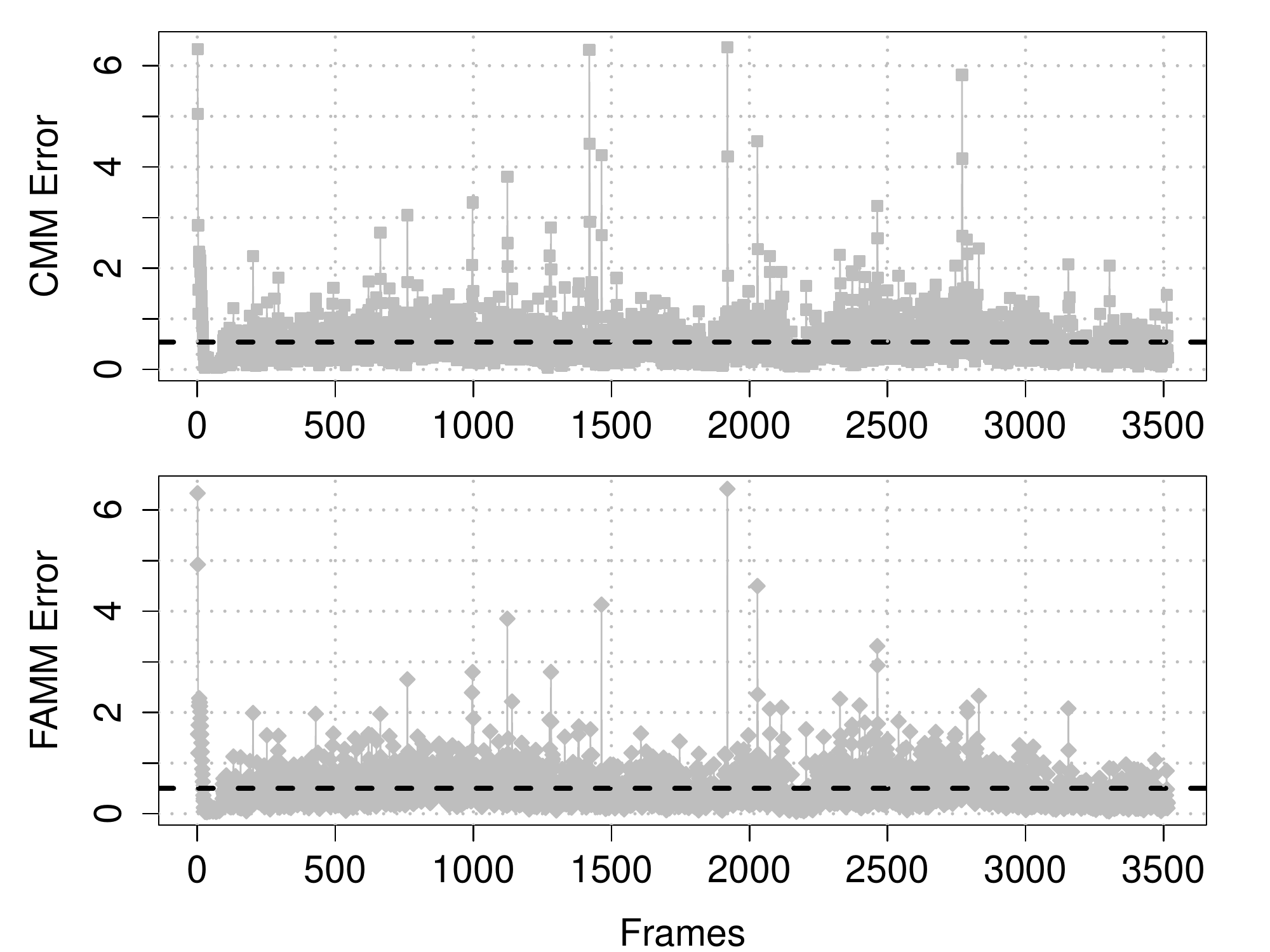}}
  \end{center}
  \caption[Filter errors for dataset 5 for CMM and FAMM]{Filter errors for dataset 5 for CMM and FAMM. The mean error is shown with the dashed line. (a) Filter error using GPS and IMU (b) Filter error using GPS, camera and IMU}
    \label{fig:filterErrorDataset5}
\end{figure}

It is also known that the state covariance matrix ($\Sigma$) is an approximation and not an actual error \cite{Reid2001}. For this reason, the filter errors are also shown in Figures~\ref{fig:filterErrorDataset1} to~\ref{fig:filterErrorDataset5}. Note that these errors are an indicator of the difference between the filter predictions and the actual values of the measurements, not the error calculation for ground-truth data, which is shown in Figures~\ref{fig:trajectoryDataset1} to~\ref{fig:trajectoryDataset5}.

Further to Figures~\ref{fig:filterErrorDataset1} to~\ref{fig:filterErrorDataset5}, a decrease in the filter error is visible when FAMM is employed to choose the motion model; see also Tables~\ref{tab:errorsGPSIMU} and~\ref{tab:errorsCameraGPSIMU}. Note that dataset 3 was captured when the tracking system was completely stationary (\emph{i.e.} no positional or rotational motion). The motion for this dataset is correctly estimated as \texttt{P0R0} which corresponds to a stationary motion model. A second thing to mention here is that the positional accuracy has been reduced to $\simeq1$ metres when the GPS is used with other sensors, where the positional accuracy only for this sensor was found as $2.5$ metres.

\begin{table*}
  \centering
  \caption[GPS and IMU filter errors]{Mean and standard deviation of the filter error for the integration of GPS and IMU. Arrows indicate an increase/decrease when the FAMM is employed.}
    \begin{tabular}{ccrrrrr}
    \toprule
          &       & \multicolumn{2}{c}{\textbf{CMM}} & \multicolumn{2}{c}{\textbf{FAMM}} &  \\
    \midrule
    \textbf{Dataset} & \textbf{Size} & \multicolumn{1}{c}{\textbf{Mean Err.}} & \multicolumn{1}{c}{\textbf{Std. Err.}} & \multicolumn{1}{c}{\textbf{Mean Err.}} & \multicolumn{1}{c}{\textbf{Std. Err.}} &  \\
    1     & 3388  & \multicolumn{1}{c}{1.106497} & \multicolumn{1}{c}{0.638612} & \multicolumn{1}{c}{1.087659} & \multicolumn{1}{c}{0.633043} & $\downarrow$ \\
    2     & 1735  & \multicolumn{1}{c}{0.884078} & \multicolumn{1}{c}{0.554315} & \multicolumn{1}{c}{0.903469} & \multicolumn{1}{c}{0.563023} & $\uparrow$ \\
    3     & 182   & \multicolumn{1}{c}{0.372074} & \multicolumn{1}{c}{0.247346} & \multicolumn{1}{c}{0.154518} & \multicolumn{1}{c}{0.232001} & $\downarrow$ \\
    4     & 1406  & \multicolumn{1}{c}{0.872035} & \multicolumn{1}{c}{0.573797} & \multicolumn{1}{c}{0.871161} & \multicolumn{1}{c}{0.569474} & $\downarrow$ \\
    5     & 3515  & \multicolumn{1}{c}{0.497994} & \multicolumn{1}{c}{0.430623} & \multicolumn{1}{c}{0.511027} & \multicolumn{1}{c}{0.447664} & $\uparrow$ \\
    \bottomrule
    \end{tabular}
  \label{tab:errorsGPSIMU}
\end{table*}

\begin{table*}
  \centering
  \caption[GPS, camera and IMU filter errors]{Mean and standard deviation of the filter error for the integration of camera,  GPS and IMU. Arrows indicate an increase/decrease when the FAMM is employed.}
    \begin{tabular}{ccrrrrr}
    \toprule
          &       & \multicolumn{2}{c}{\textbf{CMM}} & \multicolumn{2}{c}{\textbf{FAMM}} &  \\
    \midrule
    \textbf{Dataset} & \textbf{Size} & \multicolumn{1}{c}{\textbf{Mean Err.}} & \multicolumn{1}{c}{\textbf{Std. Err.}} & \multicolumn{1}{c}{\textbf{Mean Err.}} & \multicolumn{1}{c}{\textbf{Std. Err.}} &  \\
    1     & 3388  & \multicolumn{1}{c}{1.234321} & \multicolumn{1}{c}{0.885747} & \multicolumn{1}{c}{1.168871} & \multicolumn{1}{c}{0.732295} & $\downarrow$ \\
    2     & 1735  & \multicolumn{1}{c}{0.953222} & \multicolumn{1}{c}{0.576215} & \multicolumn{1}{c}{0.937526} & \multicolumn{1}{c}{0.581637} & $\downarrow$ \\
    3     & 182   & \multicolumn{1}{c}{0.155107} & \multicolumn{1}{c}{0.230386} & \multicolumn{1}{c}{0.343855} & \multicolumn{1}{c}{0.240241} & $\uparrow$ \\
    4     & 1406  & \multicolumn{1}{c}{0.894683} & \multicolumn{1}{c}{0.563316} & \multicolumn{1}{c}{0.897089} & \multicolumn{1}{c}{0.569789} & $\uparrow$ \\
    5     & 3515  & \multicolumn{1}{c}{0.543074} & \multicolumn{1}{c}{0.442076} & \multicolumn{1}{c}{0.498571} & \multicolumn{1}{c}{0.384069} & $\downarrow$ \\
    \bottomrule
    \end{tabular}
  \label{tab:errorsCameraGPSIMU}
\end{table*}

It can be seen that using an adaptive motion model (FAMM) decreases the error in general. It is also interesting to see that there are differences in errors when the fusion filter used estimates from different sensors and it can be seen that this error increased when the integration of camera, GPS and IMU is used for almost all datasets due to an additional source of noise. Note that the calculated error here is the filter error as a measure of filter convergence --not the ground truth positional error. The performance of the developed filter using three sensors in positional accuracy over the conventional GPS--IMU fusion can be seen clearly in the path plots. One exception to this is the decrease for dataset 5 when FAMM was used. For this dataset, the accuracy of the filter was also obvious from the estimated path (Figure~\ref{fig:trajectoryDataset5}(e)). For the stationary dataset (3), the error decreased when the FAMM was use in a GPS--IMU integration but increased in the case of addition of the camera estimates to these two sensors. This is suspected to be due to the large base-line requirement~\cite{Bostanci2012c} for the two-view motion estimation algorithm, which is not possible when there is no motion.

\section{Application to Outdoor Augmented Reality}
\label{sec:application}

\begin{figure}[h!t!p!b]
  \begin{center}
    \includegraphics[width=\columnwidth]{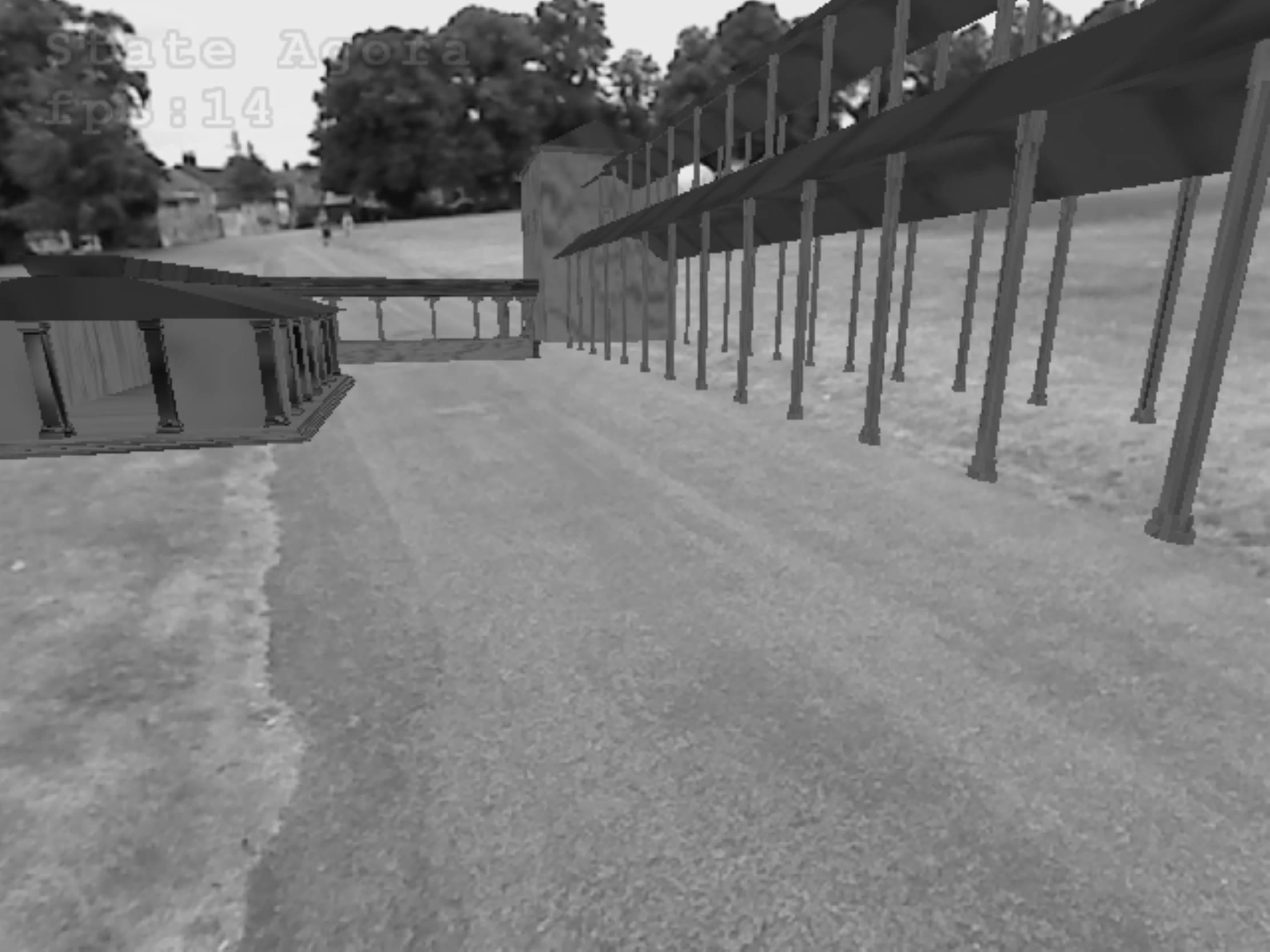}
  \end{center}
  \caption[A view from the AR application]{A view from the AR application}
  \label{fig:arApp}
\end{figure}

The application presented here allows a user walk inside a large AR model of the State \emph{Agora}, shown in Figure~\ref{fig:arApp}, in the ancient city of Ephesus, located in Turkey. The application makes use of the tracking algorithm presented in Sections~\ref{sec:sensorFusion} and~\ref{sec:fuzzyMotionModels} to allow a user's motion in real world to be reflected in the application.

\section{Conclusions}
\label{sec:conclusions}
This paper presented a user tracking system using a fusion of the motion estimates from a camera, GPS receiver and IMU sensor within a Kalman filtering framework and employing FAMM in order to reduce the filter error. A sensor fusion algorithm employing these three sensors was presented in order to overcome the tracking errors (the static and dynamic errors mentioned earlier). The filter consisted of a simple state consisting of elements for position and orientation. Initially, this filter used a simple transition function. The motion estimate from the camera was applied to the GPS position estimate and this was interpolated by the IMU estimate in order to provide a continuous and smooth navigation. The estimate for orientation was obtained using the IMU filter~\cite{Madgwick2010}. The estimations from these sensors were calculated in different threads for performance. The initial transition function was later updated by adaptive motion models which worked using fuzzy rules defining which motion model will be employed. These adaptive motion models had two parts, for the calculating the transition for the position and orientation estimates separately.

The results showed that the integration of the camera with GPS and IMU sensors provided more accurate results for tracking than a conventional GPS--IMU sensor fusion. From ~\cite{Bostanci2012c}, the vision-based algorithm was capturing the overall motion, however fine detail was missing in the motion estimate. Furthermore, motion estimation was not accurate in cases of fast movements or cases when there is no motion. For the GPS, position estimate was erroneous and not accurate. IMU was accurate for a very short term, then drift was becoming a problem.

When the three sensors were used together, these problems were significantly reduced. The motion estimates from the camera reduced the accuracy problems for the GPS. This was further improved by using the IMU so that fine detail of the motion could also be captured. This integration of several sensors solved the problems related to static errors related to the accuracy of sensors. Making use of the multiple threads allowed a better utilisation of the available resources. A second advantage of this design is that it helped reducing the dynamic errors, due to the end-to-end system delay, by providing a better frame rate.

This work also showed that multiple-motion model sensor fusion can be achieved by utilising Kalman filter innovation together with a fuzzy rule-base.  The results show that the use of fuzzy adaptive motion models can reduce the filter error and prevent divergence. It is clear that selection of the appropriate motion model depending on user's speed improves the accuracy of the fusion filter for tracking applications similar the one presented here. Future work will investigate the use of rule-reduction techniques to shrink the rule-base to set of the most commonly employed motion models.


\bibliographystyle{spmpsci}      
\bibliography{References}

\end{document}